\pdfoutput=1

\documentclass{article} %

\PassOptionsToPackage{sort&compress}{natbib}

\usepackage[final]{corl_2026} %

\usepackage{amsmath,amsfonts,bm}

\def\eqref#1{equation~\ref{#1}}

\def\1{\bm{1}}

\DeclareMathAlphabet{\mathsfit}{\encodingdefault}{\sfdefault}{m}{sl}
\SetMathAlphabet{\mathsfit}{bold}{\encodingdefault}{\sfdefault}{bx}{n}

\usepackage{hyperref}
\usepackage{url}
\usepackage{kotex}
\usepackage[final]{graphicx}       %
\usepackage{float}          %
\usepackage{amsmath}
\usepackage{amssymb}
\usepackage{pifont}
\newcommand{\cmark}{\textcolor{green!70!black}{\ding{51}}}
\newcommand{\xmark}{\textcolor{red}{\ding{55}}}
\usepackage{cleveref}
\usepackage{multirow}
\usepackage{booktabs}
\usepackage{setspace}
\usepackage{xcolor}
\usepackage{colortbl}
\usepackage{arydshln}
\usepackage{lipsum}
\usepackage{subcaption}
\usepackage[normalem]{ulem} %
\usepackage{cancel} %
\usepackage{comment} %
\usepackage{duckuments}
\usepackage{titletoc}
\usepackage{wrapfig}
\usepackage{enumitem}

\PassOptionsToPackage{colorlinks=true, linkcolor=blue, citecolor=blue, urlcolor=blue, pagebackref=true}{hyperref}
\definecolor[named]{ACMDarkBlue}{cmyk}{1,0.58,0,0.21}

\definecolor{darkgreen}{RGB}{25,200,25}
\hypersetup{colorlinks,
  linkcolor=phumacolor,  %
  citecolor=phumacolor,  %
  urlcolor=phumacolor,   %
  filecolor=phumacolor   %
}

\definecolor{ab_best}{RGB}{168, 192, 251}     %
\definecolor{ab_better}{RGB}{193, 210, 251}   %
\definecolor{ab_good}{RGB}{229, 234, 251}     %
\definecolor{ab_bad}{RGB}{251, 233, 233}
\definecolor{ab_worse}{RGB}{245, 201, 201}
\definecolor{ab_worst}{RGB}{239, 169, 169}

\definecolor{stationary}{RGB}{255,255,255}
\definecolor{yaw_rot}{RGB}{243,243,243}
\definecolor{height_move}{RGB}{226,226,226}
\definecolor{root_move}{RGB}{180,180,180}
\definecolor{header}{RGB}{235,235,235}
\definecolor{subheader}{HTML}{F0F0F0}
\definecolor{main}{RGB}{219,230,231}
\definecolor{phumacolor}{RGB}{115,144,139} %

\newlength{\cellw}

\newcommand{\datasetname}{PHUMA}

\newcommand{\crcolor}{\color{orange}}
\renewcommand{\crcolor}{}
\DeclareRobustCommand{\crnew}[1]{{\crcolor#1}} %
\makeatletter
\newif\ifcrfloat
\DeclareCaptionFont{crcolor}{\crcolor}
\g@addto@macro\@floatboxreset{\ifcrfloat\crcolor\captionsetup{font+=crcolor}\fi} %
\makeatother
\newenvironment{crblock}{\crfloattrue\crcolor}{} %

\title{\datasetname: \\ Physically Reliable Humanoid Locomotion Dataset}

\newcommand{\authorgap}{\hspace{0.7em}} %
\author{\textbf{Kyungmin Lee}$^{1}$\thanks{Equal contribution.\quad$^{\dagger}$Equal advising.}\:\:\authorgap
        \textbf{Sibeen Kim}$^{1*}$\authorgap
        \textbf{Youngdo Lee}$^{2,3}$\authorgap
        \textbf{Minho Park}$^{1}$\authorgap
        \textbf{Hyunseung Kim}$^{1,2}$ \\
        \textbf{Dongyoon Hwang}$^{1}$\authorgap
        \textbf{Donghu Kim}$^{3}$\authorgap
        \textbf{Hojoon Lee}$^{3\dagger}$\authorgap
        \textbf{Jaegul Choo}$^{1\dagger}$ \\[0.6em] %
    $^{1}$KAIST \quad $^{2}$KRAFTON \quad $^{3}$Holiday Robotics \\[0.5em] %
\href{https://davian-robotics.github.io/PHUMA/}{\textcolor{phumacolor}{\texttt{davian-robotics.github.io/PHUMA}}} \\
}

\begin{document}

\maketitle
\vspace{-1.2em} %
\begin{figure}[H]
    \centering
    \vspace{-0.2cm}
    \includegraphics[width=0.95\textwidth]{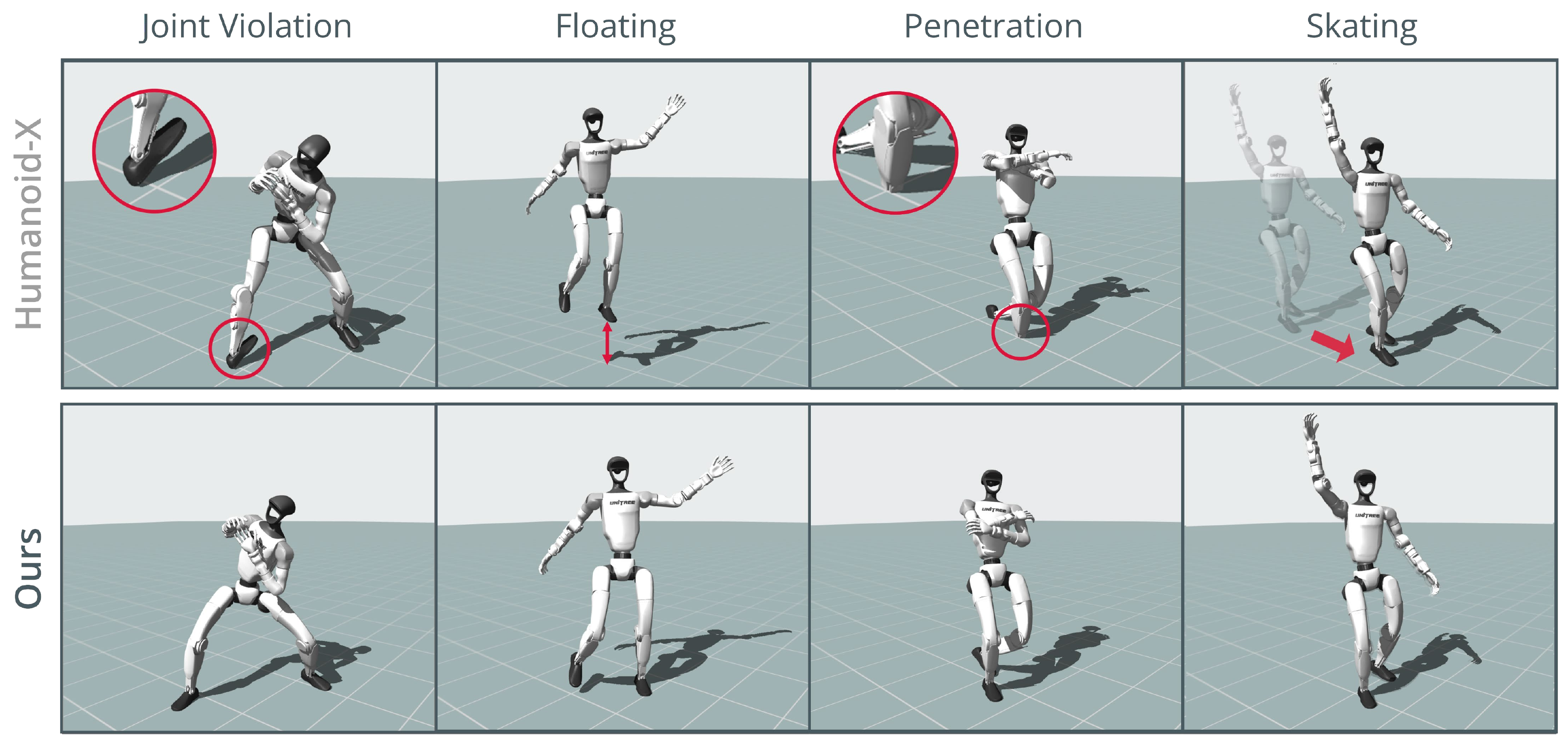}
    \caption{\textbf{Physical Reliability of Humanoid-X vs. PHUMA}. Each column illustrates four failure modes: joint violation, floating, penetration, and skating. Humanoid-X~\citep{mao2025humanoidx} (top row) often exhibits these issues due to direct video-to-motion conversion, while PHUMA (bottom row) mitigates those violations through careful data curation and physically reliable retargeting.}
    \label{fig:main_figure}
\end{figure}

\begin{abstract}
Motion imitation is a promising approach for humanoid locomotion, enabling agents to acquire humanlike behaviors. Existing methods typically rely on high-quality motion capture datasets such as AMASS, but these are scarce and expensive, limiting scalability and diversity. Recent studies attempt to scale data collection by converting large-scale internet videos, exemplified by Humanoid-X. However, they often suffer from physical artifacts such as floating, penetration, and foot skating, which hinder stable imitation.
To address this, we introduce \textbf{PHUMA}, a \textbf{P}hysically Reliable \textbf{HUMA}noid locomotion dataset produced by a two-stage pipeline combining physics-aware curation and retargeting, aggregating both motion capture and internet video into a physically reliable, 73-hour corpus. On motion tracking benchmarks, PHUMA-trained policies achieve higher success rates than those trained on AMASS and Humanoid-X, and successfully transfer zero-shot to a real Unitree G1.
\end{abstract}

\section{Introduction}
\begin{figure}[!t]
    \centering
    \vspace{-5mm}
    \includegraphics[width=1\textwidth]{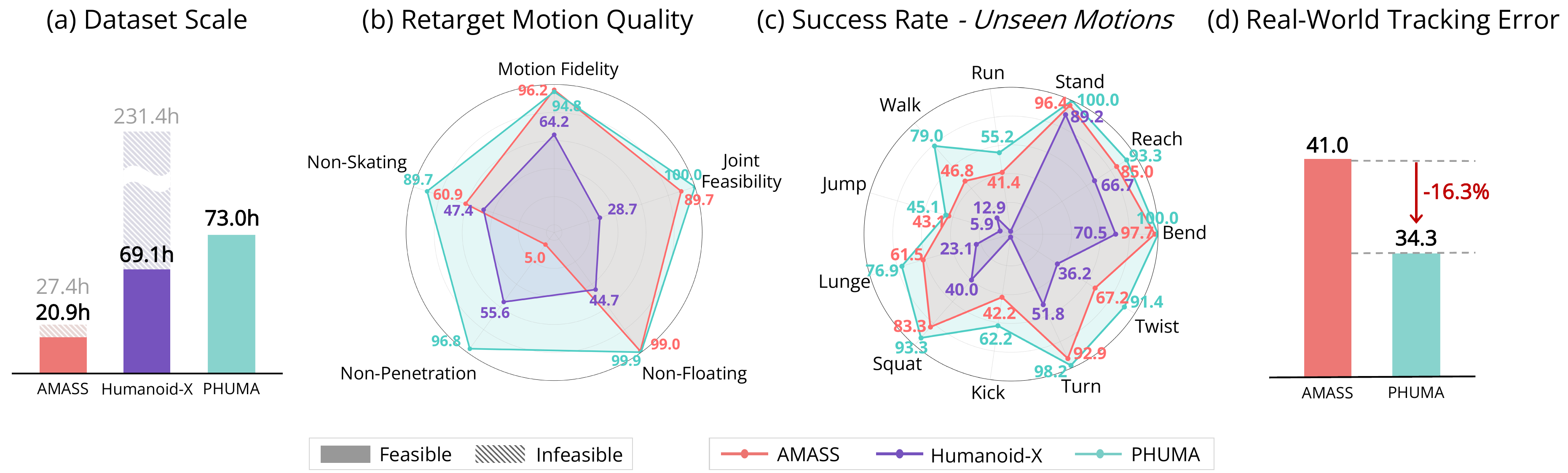}
    \vspace{-5mm}
    \caption{\textbf{Overview of datasets and performance.} 
    (a) Composition of feasible and infeasible human motion sources in each dataset. (b) Physical reliability of each dataset; AMASS is retargeted using a standard learning-based inverse kinematics method. (c) Success rate on unseen motions. (d) Motion tracking error on real Unitree G1.}
    \label{fig:figure2}
    \vspace{-5mm}
\end{figure}

\vspace{-0.1cm}

Humanoid robots are a key step toward general-purpose embodied AI, but deploying them in the real world first requires reliable and natural locomotion.
While reinforcement learning (RL) with task-oriented rewards (e.g., velocity tracking) has driven remarkable progress in quadrupedal locomotion~\citep{hwangbo2019quadloco, lee2020quadloco2, tan2018quadloco3}, applying it to humanoids often yields gaits that are effective yet non-humanlike~\citep{hansen2024tdmpc2, sferrazza2024humanoidbench}, since such rewards struggle to capture natural whole-body coordination.
To address this limitation, motion imitation has emerged as a promising paradigm, where policies are trained to replicate human movements through a three-stage pipeline: (1) collecting human motion data, (2) retargeting it to the robot's morphology, and (3) tracking the retargeted trajectories with RL~\citep{peng2018deepmimic, tessler2024maskedmimic, he2024h2o, xie2025kungfubot, he2025asap}.

Despite its promise, progress in motion imitation is fundamentally constrained by the scale, diversity, and physical feasibility of human motion data. High-quality motion capture datasets such as LaFAN1~\citep{harvey2020lafan} and AMASS~\citep{mahmood2019amass} provide a high proportion of physically feasible motions,  but are limited in scale and diversity, with content dominated by simple motions such as reaching and walking.
To overcome this scarcity, recent work has sought to scale data collection by leveraging vast internet videos.
Humanoid-X~\citep{mao2025humanoidx} exemplifies this trend by converting videos to SMPL representations~\citep{loper2023smpl, pavlakos2019expressive} using a video-to-motion model~\citep{kocabas2020vibe}, then retargeting them to humanoid embodiments. 
However, this pipeline suffers from two types of physical violations. First, the video-to-motion model often misestimates global translation, producing artifacts such as floating or ground penetration.
Second, the retargeting stage prioritizes joint alignment over physical plausibility~\citep{he2024h2o, he2024omnih2o}, leading to joint violation and foot skating as illustrated in the top row of Figure \ref{fig:main_figure}.

In response, we introduce \textbf{PHUMA}: a \textbf{P}hysically Reliable \textbf{HUMA}noid locomotion dataset that scales internet video into a high-quality training corpus by jointly filtering infeasible motions and enforcing physical constraints during retargeting.
As illustrated in Figure~\ref{fig:overview}.1, we first collect large-scale, diverse human motion data and filter out infeasible motions from Humanoid-X, such as root jitter or actions requiring external objects like sitting on chairs.
As shown in Figure~\ref{fig:overview}.2, we then apply \crnew{Physics-aware} Shape-adaptive Inverse Kinematics (PhySINK), enforcing soft joint limits, ground contact, and anti-skating constraints to eliminate violations like joint over-extension, floating, and sliding.
As a result, \datasetname~offers the most physically feasible motion among existing datasets: 73.0h total, $3.5\times$ that of AMASS and exceeding Humanoid-X's 69.1h (out of 231.4h).

To validate its effectiveness, we evaluate PHUMA across four axes: (i)-(ii) whether physics-aware curation and PhySINK retargeting contribute to downstream motion tracking, (iii) comparison against existing datasets, and (iv) zero-shot deployment on a real Unitree G1. We use the MaskedMimic~\citep{tessler2024maskedmimic} for in-simulation tests on both Unitree G1 and H1-2 humanoids, and BeyondMimic~\citep{truong2025beyondmimic} for real-world deployment on a Unitree G1. On 504 self-recorded videos across 11 motion types, policies trained with PHUMA outperform both AMASS and Humanoid-X across all motion types in success rate (Figure~\ref{fig:figure2}(c)). Furthermore, on 24 motions sampled from the PHUMA test set (6 per category: stationary, angular, vertical, and horizontal), the PHUMA-trained policy achieves $16.3\%$ lower tracking error than the AMASS-trained policy in the real world (Figure~\ref{fig:figure2}(d)). 

\section{Related Work}

\datasetname~focuses on constructing a large-scale, physically reliable humanoid locomotion dataset, requiring two components: (1) collecting diverse human motion data and (2) retargeting it to the humanoid robots.

\subsection{Human Motion Data}
\vspace{-2mm}
Human motion data, typically in the SMPL format~\citep{loper2023smpl, pavlakos2019expressive}, comes from two main sources: motion capture (mocap) and video-based reconstruction. Mocap~\citep{CMU_mocap, zhang2022egobody, al2023locomujoco} provides accurate kinematics but demands expensive instrumentation and studio effort. LaFAN1~\citep{harvey2020lafan}, for instance, offers high fidelity but only a few hours of motion, while AMASS~\citep{mahmood2019amass} unifies many mocap datasets yet remains walking-centric, limiting diversity. Concurrent work such as Bones-SEED~\citep{bonesseed} scales open mocap data through commercial studio production.
Driven by advances in video-to-motion recovery~\citep{kocabas2020vibe, shin2024wham, wang2024tram, shen2024gvhmr, gu2025humanoid}, human video has emerged as a scalable, diverse alternative. Recent works~\citep{lin2023motion, zhang2025motion, chung2021haa500, cai2022humman, tsuchida2019aist, fan2025go, ni2025generated, wei2025unveiling, wang2026humanx, weng2025hdmi} leverage this scalability and diversity, with Humanoid-X~\citep{mao2025humanoidx} notably providing abundant data from the Internet. However, video-derived motions often exhibit severe jitter across frames and physical artifacts such as foot sliding or ground penetration~\citep{luo2023phc, luo2024universal, goel2023humans, ye2023decoupling, yu2021human, ugrinovic2024multiphys}. 
To mitigate these artifacts, recent methods employ automated filtering, either by discarding motions a pretrained tracking policy cannot reproduce in simulation~\citep{he2025asap}, or by detecting foot-ground contacts through hand-tuned thresholds (e.g., zero-velocity and ankle-height)~\citep{xie2025kungfubot}. Yet, the former risks rejecting feasible but out-of-distribution motions, and the latter misjudges contact when video-estimated global translation is noisy.
\crnew{We aggregate large-scale motion data from both mocap and human video, and instead curate it with physics-aware criteria computed directly from the motion, without relying on a tracking policy. To keep contact detection reliable, we measure contact against a robust ground plane estimated by majority voting, and filter out clips that are jittery, unstable, or floating.}

\subsection{Humanoid Motion Retargeting}
\vspace{-2mm}
Building on its success in physics-based character control~\citep{peng2018deepmimic, wagener2022mocapact, luo2021dynamics, luo2023phc, hansen2025hierarchical, tessler2024maskedmimic, tirinzoni2025zeroshot, zhang2025physics, zhang2026script}, human motion data is increasingly adapted for humanoid robotics~\citep{radosavovic2024learning, fu2024humanplus, cheng2024exbody, ji2024exbody2, chen2025gmt, xie2025kungfubot, truong2025beyondmimic, li2025clone, luo2025sonic, li2026omnitrack, li2025amo, wang2026omnixtreme, rempe2026kimodo, zhao2025resmimic, yin2025visualmimic}. This requires motion retargeting~\citep{gleicher1998ik}, mapping human motions onto robots that share a humanoid topology but differ in kinematics and body proportions~\citep{kim2025pyroki, ho2010spatial, zhang2023skinned}.
Standard Inverse Kinematics (IK) methods~\citep{radosavovic2024humanoid, pink, ze2025twist} are widely adopted for their simplicity, but they often ignore these morphological differences, leading to unnatural artifacts such as misaligned foot orientations. While approaches like GMR~\citep{araujo2025gmr} demonstrate that careful engineering can improve IK outcomes, they still rely heavily on heuristic scale adjustments and remain prone to floating artifacts. To explicitly address shape discrepancies, Shape-adaptive Inverse Kinematics (SINK) methods~\citep{he2024h2o, he2024omnih2o, he2025hover, cheynel2023sparse, allshire2025visual} adapts the human body shape to the robot before pose alignment, which is effective at pose matching but physically under-constrained.
To enforce physical realism, several works~\citep{chen2025whole, yang2025omniretarget, pan2025spider, zhang2026kinodynamic, wang2026spark, muller2026reactor} add contact constraints, but assume clean motion where the ground follows from the lowest foot position; on noisy video-derived motion this collapses, flagging nearly every frame as floating. Neural retargeting~\citep{chen2025implicit, zhao2026make} learns the mapping from data, \crnew{and dynamics-aware methods such as SPIDER~\citep{pan2025spider} refine motions with sampling-based optimization,} yet \crnew{both} still \crnew{need} a prior retargeting method, for training pairs \crnew{and for initialization, respectively}.
\crnew{We retarget the curated human motion data with Physics-aware SINK} (PhySINK), which augments SINK with joint feasibility, grounding, and anti-skating terms\crnew{ and takes contact signals from the ground plane estimated during curation, keeping them reliable on noisy video}.

\section{Method}

\begin{figure}[t!]
    \centering
    \includegraphics[width=0.99\textwidth]{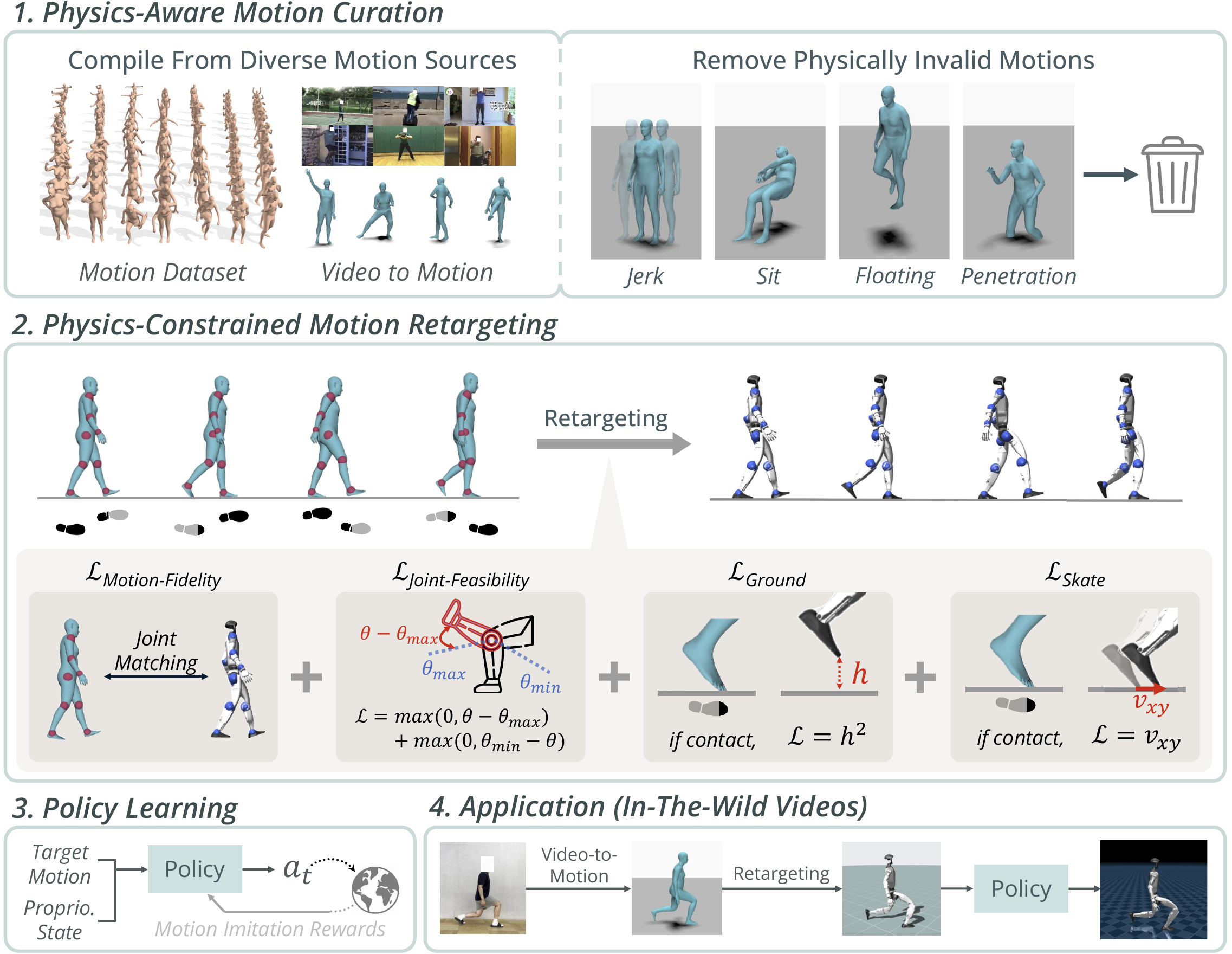}
    \caption{\textbf{Overview of the \datasetname~pipeline.} Our four-stage pipeline for motion imitation learning includes: (1) Motion Curation, where we filter out problematic motions from a diverse dataset; (2) Motion Retargeting, where the filtered motions are retargeted to the humanoid using PhySINK; (3) Policy Learning, where a policy is trained to imitate the retargeted motions; and (4) Inference, where the trained policy is used to control the humanoid, enabling it to imitate motions from unseen videos processed by a video-to-motion model.}
    \label{fig:overview}
    \vspace{-5mm}
\end{figure}

Our goal is to construct \datasetname, a large-scale, physically reliable dataset for humanoid locomotion. We build upon the Humanoid-X motions~\citep{mao2025humanoidx}, which are rich in scale but exhibit physical artifacts. We first apply physics-aware curation to filter out problematic motions (Section~\ref{subsec:method_curation}). Next, to address artifacts introduced during the retargeting process itself, we employ PhySINK, our \crnew{physics-aware} retargeting method that adapts the curated motion to the humanoid while enforcing physical reliability (Section~\ref{subsec:method_retargeting}). Our full pipeline is illustrated in Figure~\ref{fig:overview}.

\subsection{Physics-Aware Motion Curation}
\label{subsec:method_curation}
\vspace{-2mm}
Our curation pipeline addresses three key artifacts in raw motion data: severe jitter, physical instability from interactions with objects absent in the humanoid's environment (e.g., chairs), and incorrect foot-ground contact (e.g., floating or penetration).

\textbf{Jitter and instability.} To mitigate high-frequency jitter, we apply a low-pass Butterworth filter, which smooths the motion by removing high-frequency components while preserving the overall trajectory (Appendix~\ref{appendix:lpf}). To detect physical instability, such as sitting on a non-existent chair, we compute the center-of-mass (CoM) distance from the base of support, defined as the ground area covered by the feet in contact with the floor.

\textbf{Ground contact.} To ensure that the feet make proper contact with the ground, a fixed ground plane is required as a reference. However, recovered motions are often defined in a camera's coordinate frame, which has no fixed ground, causing floating and penetration. To address this, we estimate a global ground plane by majority voting over the heights of vertices on the foot meshes, selecting the height most consistent with ground contact. Using this estimate, we subtract the ground plane's height from the z-coordinate of every joint in every frame, shifting the entire motion so that the ground plane aligns with $z = 0$ (Appendix~\ref{appendix:ground_contact}). We then compute a contact score for each foot region (e.g., heel, toe), measuring how close its vertices are to the ground plane in each frame.

With a reliable ground plane established, we segment all sequences into 4-second clips and discard any clip that exhibits (i) excessive jerk, indicating high-frequency jitter; (ii) a CoM position far outside its base of support, indicating physical instability; or (iii) insufficient foot-ground contact, indicating floating. By filtering at the clip level rather than the full sequence, we preserve valid segments even when parts of the original sequence are flawed (Appendix~\ref{appendix:filtering}).

Finally, we augment these curated motions with data from LaFAN1, LocoMuJoCo, and our own video captures.
The resulting \datasetname~dataset is a large-scale collection of 73.0 hours of physically reliable motion across 76.0K clips, with the full composition in Appendix~\ref{appendix:dataset_composition}.

\subsection{\texorpdfstring{\crnew{Physics-Aware}}{Physics-Aware} Motion Retargeting}
\label{subsec:method_retargeting}
After curation, the next step is to retarget human motions to the humanoid embodiment. Standard inverse kinematics \textbf{(IK)} rigidly maps human joint positions to the humanoid skeleton, distorting the original motion when the two skeletons differ in proportion. Shape-adaptive inverse kinematics \textbf{(SINK)} preserves motion style by first reshaping the human body to match the humanoid's kinematic structure, but does not account for physical plausibility, leading to artifacts such as joint violations and unrealistic ground interactions.

\textbf{PhySINK.} We propose \crnew{Physics-aware} Shape-adaptive Inverse Kinematics (PhySINK), which augments SINK with three physics-aware constraints. Our key insight is that physical artifacts in retargeting arise from three distinct sources: the mechanical limits of the robot, the ground plane, and contact consistency. Each can be addressed by a dedicated loss term\crnew{, added to a motion fidelity loss} during optimization.

\begin{crblock}
\textbf{Motion Fidelity Loss.} We optimize the humanoid joint positions $q_t$ and root translation $\gamma_t$ over frames $t$ so that the retargeted motion closely matches the SMPL-X motion. The loss combines global position matching, local joint matching, and temporal smoothness:
\begin{equation}
    \mathcal{L}_\text{global-match} = \sum_{t}\sum_{i} \left\| p_{i}^\text{SMPL-X}(t) - p_{i}^\text{Humanoid}(t) \right\|_1
\end{equation}
\begin{align}
    \mathcal{L}_\text{local-match} &= \sum_{t}\sum_{i \ne j} m_{ij} \underbrace{\left\| \Delta p_{ij}^\text{SMPL-X}(t) - \Delta p_{ij}^\text{Humanoid}(t) \right\|_2^2}_{\text{position}} \\
    \nonumber &+ \sum_{t}\sum_{i \ne j} m_{ij} \underbrace{\left( 1 - \left\langle \Delta p_{ij}^\text{SMPL-X}(t),\ \Delta p_{ij}^\text{Humanoid}(t) \right\rangle \right)}_{\text{orientation}}
\end{align}
\begin{equation}
    \mathcal{L}_\text{smooth} = \sum_{t}\bigl\lVert \dot{q}_{t}-2\dot{q}_{t+1}+\dot{q}_{t+2}\bigr\rVert_1 + \sum_{t}\bigl\lVert \dot{\gamma}_{t}-2\dot{\gamma}_{t+1}+\dot{\gamma}_{t+2}\bigr\rVert_1
\end{equation}
\begin{equation}
    \mathcal{L}_\text{Fidelity} = w_\text{global-match}\mathcal{L}_\text{global-match} + w_\text{local-match}\mathcal{L}_\text{local-match} + w_\text{smooth}\mathcal{L}_\text{smooth}
\end{equation}
where $p_{i}(t)$ is the global 3D position of joint $i$ at frame $t$, $\Delta p_{ij}(t)$ is the displacement between joints $i$ and $j$, and $m_{ij}$ is 1 if $i$ and $j$ are immediate neighbors in the kinematic tree and 0 otherwise. Here $\dot{q}_t$ and $\dot{\gamma}_t$ denote the joint and root velocities, and each $w$ is a scalar weight.

\textbf{Joint Feasibility Loss.} We penalize joint angles $q_t$ and velocities $\dot{q}_t$ that exceed 98\% of the robot's mechanical limits, preventing over-extension and unrealistic actuation:
\begin{align}
    \mathcal{L}_{\text{pos-violation}} &= \sum_{t} \left[ \max(0, q_t - 0.98q_{\text{max}}) + \max(0, 0.98q_{\text{min}} - q_t) \right] \\
    \mathcal{L}_{\text{vel-violation}} &= \sum_{t} \left[ \max(0, \dot{q}_t - 0.98\dot{q}_{\text{max}}) + \max(0, 0.98\dot{q}_{\text{min}} - \dot{q}_t) \right] \\
    \mathcal{L}_\text{Feasibility} &= \mathcal{L}_{\text{pos-violation}} + \mathcal{L}_{\text{vel-violation}}
\end{align}
where $[q_\text{min}, q_\text{max}]$ and $[\dot{q}_\text{min}, \dot{q}_\text{max}]$ are the position and velocity limits of each joint. Penalizing values beyond 98\% of each limit leaves a small margin below the mechanical limits of the robot.

\textbf{Grounding Loss.} We align the foot height with the ground plane during contact, eliminating floating and penetration:
\begin{equation}
    \mathcal{L}_\text{Ground} = \sum_{i \in \{\text{LH}, \text{LT}, \text{RH}, \text{RT}\}}\sum_{t} c_t^{i}\left\| p_{t}^{i}(z)\right\|_2^2
\end{equation}
where $c_t^i$ is the ground-contact ratio of foot region $i$ (left heel, left toe, right heel, right toe) at frame $t$, obtained during curation (Appendix~\ref{appendix:filtering}), and $p_t^i(z)$ is its height. 

\end{crblock}
\begin{table}[t!]
    \centering
    \vspace{-1mm}
    \caption{\textbf{Quantitative comparison and ablation study of retargeting methods.} We evaluate performance on the G1 humanoid, showing the progressive impact of adding each of our proposed physical constraint losses. (H1-2 results in Appendix~\ref{appendix:h12_results})}
    \vspace{2mm}
    \label{tab:retarget_ablation}
    \scalebox{0.71}{
        \begin{tabular}{@{}lcccccccc@{}}
        \toprule
         & Motion Fidelity (\%)  & Joint Feasibility (\%) & Non-Floating (\%) & Non-Penetration (\%) & Non-Skating (\%) \\
        \midrule
        IK                                & 27.6          & 91.7           & 55.6          & 47.8          & 59.7          \\
        GMR & 56.3 & 81.8 & 14.7 & \textbf{100.0} & 67.7 \\
        SINK                              & 94.8          & 95.9           & 96.4          & 14.9          & 55.4          \\
        + Joint Feasibility Loss          & \textbf{94.9} & \textbf{100.0} & 96.4          & 14.8          & 55.6          \\
        + Grounding Loss                  & \textbf{94.9} & \textbf{100.0} & \textbf{99.9} & 97.2          & 53.6          \\
        + Skating Loss = PhySINK          & 94.8          & \textbf{100.0} & \textbf{99.9} & 96.8          & \textbf{89.7}          \\

        \bottomrule
        \end{tabular}
    }
    \vspace{-2mm}
\end{table}

\begin{crblock}
\textbf{Skating Loss.} We suppress horizontal foot velocity during contact, preventing the foot from sliding:
\begin{equation}
    \mathcal{L}_\text{Skate} = \sum_{i \in \{\text{LH}, \text{LT}, \text{RH}, \text{RT}\}}\sum_{t} c_t^{i}\left\| \dot{p}_{t}^{i}(x,y)\right\|_2
\end{equation}
where $\dot{p}_t^i(x,y)$ is the horizontal velocity of region $i$ at frame $t$.

\end{crblock}
The full objective is:
\begin{equation}
    \label{eq:PhySINK}
    \mathcal{L}_\text{PhySINK} = \mathcal{L}_\text{Fidelity} + w_\text{Feasibility}\mathcal{L}_\text{Feasibility} + w_\text{Ground}\mathcal{L}_\text{Ground} + w_\text{Skate}\mathcal{L}_\text{Skate}
\end{equation}
By jointly optimizing these terms, PhySINK produces motions that faithfully reflect the human motion while remaining physically reliable.

\textbf{Evaluation.} We measure physical reliability across five complementary metrics, each defined as the percentage of frames satisfying a physical criterion: \textit{Motion Fidelity} (per-joint position error $<$ 10\,cm and per-link orientation error $<$10°), \textit{Joint Feasibility} (joint positions and velocities within 98\% of mechanical limits), \textit{Non-Floating} (foot within 1\,cm above ground during contact), \textit{Non-Penetration} (foot within 1\,cm below ground during contact), and \textit{Non-Skating} (horizontal foot velocity $<$ 10\,cm/s during contact). We retarget curated motions onto the Unitree G1 and H1-2~\citep{UnitreeG1, UnitreeH12}, comparing against a standard IK solver~\citep{Zakka2025mink}, GMR~\citep{araujo2025gmr}, and SINK.

As shown in Table~\ref{tab:retarget_ablation}, IK and GMR rely on linear scaling between the human and robot skeletons, which fails to account for body shape differences and loses motion fidelity. SINK recovers fidelity by adapting body shape, but suffers from physical violations. GMR achieves the highest Non-Penetration by lowering the entire motion to its lowest contact point, but this causes severe floating elsewhere. PhySINK is the only method that performs consistently well across all five metrics. Qualitative comparisons are provided in Appendix~\ref{appendix:qualitaive_retarget}.

\section{Experiments}
In this section, we evaluate the effectiveness of our curation and retargeting pipeline, together with the resulting PHUMA dataset, along four axes corresponding to the following research questions:

\vspace{-0.5mm}
\textbf{RQ1.}  Does physics-aware data curation improve downstream motion imitation?

\vspace{-1mm}
\textbf{RQ2.} How does our proposed PhySINK retargeting method compare with established retargeting approaches (IK, GMR, SINK) in terms of motion imitation performance?

\vspace{-1mm}
\textbf{RQ3.} How effective is PHUMA as a training data for motion imitation, compared with prior humanoid motion datasets (LaFAN1, AMASS, Humanoid-X)?

\vspace{-1mm}
\textbf{RQ4.} Does PHUMA's advantage in motion tracking extend from simulation to the real robot?

\subsection{Experiment Setup}
\label{experiment_setup}
\textbf{Training.}
The motion tracking framework for in-simulation evaluation is based on MaskedMimic~\citep{tessler2024maskedmimic}, while BeyondMimic~\citep{truong2025beyondmimic} is utilized for sim-to-real transfer. 
In both approaches, policies are trained to imitate retargeted motions by maximizing motion tracking reward signals via PPO~\citep{schulman2017ppo}.
In-simulation experiments involve both the Unitree G1 (29 DoF) and H1-2 (21 DoF, excluding wrist joints) humanoids, whereas real-world validation is performed exclusively on the Unitree G1. Comprehensive training details, including the observation space, reward function, and PPO parameters, are provided in Appendix~\ref{appendix:details_imitation_learning}.

\textbf{Evaluation.}
Performance evaluation of the trained policies is conducted across two distinct datasets. The first comprises the PHUMA test split, containing approximately 7.5K motions (10\% of PHUMA) held out during training. The second consists of 504 self-collected video sequences, which are converted into motion trajectories via a video-to-motion model; processing details are provided in Appendix~\ref{appendix:self_collected_video}.
To address RQ1--RQ3, the success rate~\citep{luo2023phc, he2024h2o, he2025asap, xie2025kungfubot} is measured using a 0.15m threshold, which is stricter than the conventional 0.5m limit (see Appendix~\ref{appendix:performance_threshold}). For RQ4, due to the unavailability of mocap equipment in the real-world setup---precluding the recording of global translations---the evaluation relies on local mean per-joint position error (MPJPE, mm) alongside DoF velocity (deg/frame) and acceleration (deg/frame$^2$) errors instead of their global counterparts. 

\subsection{Effect of Physics-Aware Curation }

To isolate the contribution of physics-aware curation, we conduct this evaluation under a controlled setup. We retarget all motions with SINK, which applies no physical correction at the retargeting stage, so that any change in downstream performance can be attributed to curation alone. LaFAN1 and LocoMuJoCo are excluded, as they are already provided in retargeted form and would entangle retargeting \crnew{and} curation effects, and the validation split is held out throughout.
As shown in \crnew{Table}~\ref{tab:curation_performance}, applying any single filtering component improves motion tracking performance over the no-curation baseline, and applying all components yields a substantially larger gain, even though the fully filtered set contains roughly one quarter of the unfiltered motion volume. These results indicate that physics-aware curation contributes meaningfully to motion tracking performance, with data quality outweighing raw quantity. \crnew{A leave-one-out ablation further supports this, removing any single filter lowers performance even though it increases the data volume, with the foot contact filter contributing the most (Appendix~\ref{appendix:loo_curation}).}

\begin{table}[h]
      \centering
      \vspace{-2mm}
      \caption{\textbf{Imitation Performance: Effect of Physics-Aware Curation.}
      MaskedMimic policies trained on Humanoid-X subsets with varying curation filters,
      on Unitree G1. All variants use SINK retargeting to isolate filtering effect.
      Hours: remaining dataset size after filtering.}
      \label{tab:curation_performance}
      \vspace{2mm}
      \scalebox{0.71}{
          \begin{tabular}{l | c | cccc | ccccc | ccccc}
              \toprule
              & &
              \multicolumn{4}{c|}{Curation Filter} &
              \multicolumn{5}{c|}{\datasetname\ Test} &
              \multicolumn{5}{c}{Unseen Video} \\
              \cmidrule(lr){3-6}
              \cmidrule(lr){7-11}
              \cmidrule(lr){12-16}
              Variant & Hours & Jerk & FC & Height & BoS
              & Total & Stat. & Ang. & Vert. & Horiz.
              & Total & Stat. & Ang. & Vert. & Horiz. \\
              \midrule
              \rowcolor{gray!15}
              Humanoid-X     & 231 & \xmark & \xmark & \xmark & \xmark & 50.6 & 78.4 & 43.0 & 26.0 & 31.8 & 39.1 & 78.0 & 39.6 & 23.0 & 6.5 \\
                  + Jerk only          & 141 & \cmark & \xmark & \xmark & \xmark & 83.0 & 91.5 & 77.6 & 72.3 & 88.5 & 77.6 & \textbf{100.0} & 78.8 & \textbf{62.8} & 61.5 \\
              \rowcolor{gray!15}
              + FC only            & 123 & \xmark & \cmark & \xmark & \xmark & 82.9 & 91.3 & 77.8 & 70.4 & 89.1 & 78.2 & 99.1 & 79.3 & 61.7 & 65.9 \\
              + Height only        & 136 & \xmark & \xmark & \cmark & \xmark & 80.6 & 90.7 & 74.6 & 69.3 & 85.1 & 74.0 & 98.3 & 76.4 & 61.7 & 50.5 \\
              \rowcolor{gray!15}
              + BoS only           & 110 & \xmark & \xmark & \xmark & \cmark & 81.3 & 90.2 & 76.2 & 72.0 & 84.4 & 74.8 & 96.6 & 79.8 & 54.3 & 57.1 \\
              \textbf{All filters} & 62  & \cmark & \cmark & \cmark & \cmark & \textbf{87.0} & \textbf{93.0} & \textbf{82.7} & \textbf{81.8} & \textbf{90.3} & \textbf{79.6} & 98.3 & \textbf{82.8} & 59.6 & \textbf{69.2} \\
              \bottomrule
          \end{tabular}
      }
  \end{table}

\subsection{PhySINK Retargeting Method Effectiveness}
\label{subsec:exp_retarget}
To evaluate PhySINK, we compare it against three baselines (IK, GMR, and SINK) using the AMASS dataset. Since AMASS consists of clean, motion-capture data, it effectively removes the influence of data curation, allowing us to isolate the retargeting performance. We retarget the identical motions with each method and train a separate full-state tracking policy for each baseline.
As shown in Table~\ref{tab:tracking_comparison_on_retarget}, PhySINK outperforms the baselines on both evaluation sets, demonstrating that physically reliable retargeting leads to better imitation performance. Notably, the largest gains appear on angular and vertical motions, where balance is critical and physical reliability of training data directly affects policy learning. \crnew{This advantage also persists beyond the retargeting distribution of the evaluation references, as PhySINK still achieves the highest success rate when the same motions are retargeted with GMR (Appendix~\ref{appendix:gmr_reference}).}

\begin{table}[h]
    \centering
    \vspace{-3mm}
    \caption{\textbf{Motion tracking performance across retargeting approaches.} Success rates of policies trained on AMASS data retargeted by IK, GMR, SINK, and PhySINK, evaluated on the G1 humanoid robot using two test sets: PHUMA Test and Unseen Video. (H1-2 results in Appendix~\ref{appendix:h12_results})}
\vspace{2mm}
    \label{tab:tracking_comparison_on_retarget}
    \scalebox{0.77}{
        \begin{tabular}{l*{11}{c}}
            \toprule
            &  \multicolumn{5}{c}{\datasetname \ Test} & \multicolumn{5}{c}{Unseen Video} \\
            \cmidrule(lr){2-6}
            \cmidrule(lr){7-11}
            Retarget  & Total & Stationary & Angular & Vertical & Horizontal & Total & Stationary & Angular & Vertical & Horizontal  \\
            \midrule
            IK                & 52.8 & 75.3 & 43.9 & 24.3 & 44.2 & 54.6 & 89.3 &
54.6 & 32.7 & 43.3 \\
            GMR              & 61.7 & 80.2 & 51.7 & 32.5 & \textbf{74.0} & 59.3 & 87.1 & 64.5 & 31.9 & 40.7 \\            
            SINK             & 76.2 & 88.5 & 72.1 & 56.8 & 66.8 & 70.2 & 90.7 &
75.0 & 62.7 & 44.1 \\
            \textbf{PhySINK} & \textbf{79.5} & \textbf{89.9} & \textbf{76.1} &
\textbf{61.1} & 69.5 & \textbf{72.8} & \textbf{93.3} & \textbf{78.2} &
\textbf{65.6} & \textbf{47.3} \\

            \bottomrule
        \end{tabular}
    }

    \vspace{-5mm}
\end{table}

\subsection{PHUMA Dataset Effectiveness}
Having demonstrated PhySINK's effectiveness, we now compare \datasetname~against existing humanoid datasets. We train full-state policies on four datasets with different characteristics: LaFAN1 (small-scale, high-quality), AMASS (medium-scale, moderate-quality), Humanoid-X (large-scale, lower-quality), and \datasetname~(large-scale, high-quality). For AMASS, we apply the widely-used SINK retargeting method since it provides human motion source data, while LaFAN1 and Humanoid-X are used directly as pre-existing humanoid datasets.
As shown in Table~\ref{tab:tracking_comparison_on_dataset}, policies trained on \datasetname~achieve the highest success rates across all motion categories and both humanoids. The results reveal that neither scale nor quality alone is sufficient: Humanoid-X, despite its large size, underperforms due to quality issues, while the cleaner LaFAN1 and AMASS lack coverage in several motion types. By combining large scale with high-quality motions, \datasetname~delivers consistently superior performance across diverse behaviors. \crnew{As with PhySINK, this advantage holds when the evaluation motions are retargeted with GMR (Appendix~\ref{appendix:gmr_reference}). \datasetname~also outperforms AMASS retargeted with PhySINK in this setting, indicating that the gain comes from the data itself and not from PhySINK alone.}

\begin{table}[h]
    \centering
    \vspace{-3mm}
    \caption{\textbf{Motion tracking performance across datasets.} Success rates of policies trained on LaFAN1, AMASS, Humanoid-X, and PHUMA, evaluated across motion categories on the G1 humanoid robot using two test sets: PHUMA Test and Unseen Video. (H1-2 results in Appendix~\ref{appendix:h12_results})}
    \vspace{0mm}
    \label{tab:tracking_comparison_on_dataset}
    \scalebox{0.7}{
        \begin{tabular}{l*{11}{c}}
            \toprule
            & & \multicolumn{5}{c}{\datasetname \ Test} & \multicolumn{5}{c}{Unseen Video} \\
            \cmidrule(lr){3-7}
            \cmidrule(lr){8-12}
            Dataset & Hours & Total & Stationary & Angular & Vertical & Horizontal & Total & Stationary & Angular & Vertical & Horizontal  \\
            \midrule
            LaFAN1                & 2.4   & 46.1 & 66.1& 36.2 & 24.0 & 42.5 & 28.4 & 46.9 & 28.4 & 19.6 & 10.5 \\
            AMASS                 &20.9 & 76.2 & 88.5 & 72.1 & 56.8 & 66.8 &
78.2 &90.7 & 75.0 & 62.7 & 44.1 \\
            Humanoid-X            & 231.4 & 50.6 & 78.4 & 43.0 & 26.0 & 31.8 &
39.1 & 78.0 & 39.6 & 23.0 & 6.5 \\
            \textbf{\datasetname} & 73.0  & \textbf{92.7} & \textbf{95.6} & \textbf{91.7} & \textbf{86.0} & \textbf{85.6} & \textbf{82.9} & \textbf{96.7} &  \textbf{88.0} & \textbf{71.8} & \textbf{67.1} \\

            \bottomrule
        \end{tabular}
    }

    \vspace{-2mm}
\end{table}

\subsection{Real-World Experiment}

To verify whether the advantage of PHUMA-trained policies transfers to physical hardware, we adopt BeyondMimic~\citep{truong2025beyondmimic}, a motion tracking framework with proven sim-to-real transfer on a real Unitree G1. Since the original BeyondMimic is designed for single-motion tracking, we extend it to multi-motion tracking by augmenting the policy's observation space with a learned latent encoding of the future reference trajectory; full details are provided in Appendix~\ref{appendix:details_imitation_learning}. We train policies on AMASS and PHUMA in IsaacSim and evaluate zero-shot tracking performance on 6 motions per category (stationary, angular, vertical, and horizontal) in the PHUMA test split.

\begin{figure}[H]
    \centering
    \includegraphics[width=1\textwidth]{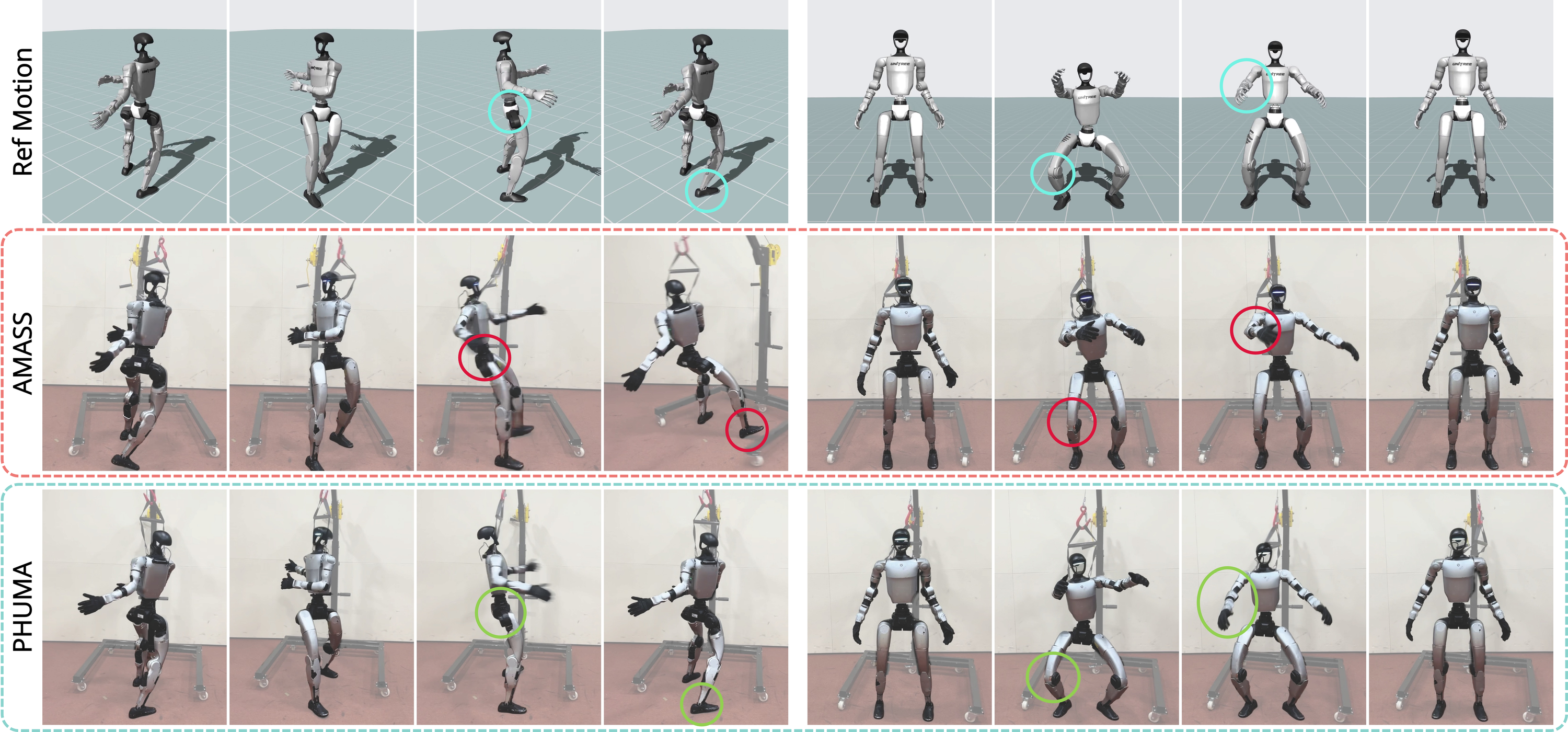}
    \caption{\textbf{Real-World Deployment}. Each row shows the reference motion (top), the motion tracked by the AMASS-trained policy (middle), and the motion tracked by the PHUMA-trained policy (bottom), demonstrating that PHUMA achieves closer alignment with the reference across all sequences. }
    \label{fig:figure_real}
    \vspace{-2mm}
\end{figure}

As shown in Table~\ref{tab:tracking_metrics_sim_to_real} and Figure~\ref{fig:figure_real}, the PHUMA-trained policy achieves lower tracking errors across all metrics and exhibits qualitatively more faithful motion tracking, indicating that PHUMA's advantages in scale and physical reliability extend from simulation to the real robot.

\begin{table}[h]
      \centering
      \caption{\textbf{Real-World Tracking Results.} Tracking errors of policies trained on AMASS and PHUMA, evaluated on 6 motions per category from the PHUMA test set, on the Unitree G1.}
      \label{tab:tracking_metrics_sim_to_real}
      \vspace{0mm}
      \scalebox{0.76}{
          \begin{tabular}{l*{15}{c}}
              \toprule
              & \multicolumn{5}{c}{$E_{\mathrm{mpjpe}}$ (mm) $\downarrow$} & \multicolumn{5}{c}{$E_{\mathrm{dof\_vel}}$ (deg/frame) $\downarrow$} & \multicolumn{5}{c}{$E_{\mathrm{dof\_acc}}$ (deg/frame$^2$) $\downarrow$} \\
              \cmidrule(lr){2-6}
              \cmidrule(lr){7-11}
              \cmidrule(lr){12-16}
              Dataset & Total & Stat. & Ang. & Vert. & Hor. & Total & Stat. & Ang. & Vert. & Hor. & Total & Stat. & Ang. & Vert. & Hor. \\
              \midrule
              AMASS                 & 41.0 & 27.3 & 47.5 & 49.9 & 39.4 & 3.20 & 1.81 & 4.10 & 3.14 & 3.77 & 1.50 & 0.77 & 2.09 & 1.33 & \textbf{1.80} \\
              \textbf{\datasetname} & \textbf{34.3} & \textbf{23.2} & \textbf{40.4} & \textbf{42.6} & \textbf{31.1} & \textbf{2.69} & \textbf{1.41} & \textbf{3.07} & \textbf{2.68} & \textbf{3.60} & \textbf{1.29} & \textbf{0.58} & \textbf{1.43} & \textbf{1.18} & 1.96 \\
              \bottomrule
          \end{tabular}
      }
      \vspace{-2mm}
  \end{table}

\section{Limitations}

While PHUMA advances the scale and physical reliability of humanoid motion data, several directions remain open for future work. First, our pipeline focuses on flat-terrain locomotion and does not yet cover motions involving scene interaction (e.g., object manipulation, sitting, climbing) or uneven terrain; extending it to such settings would require jointly modeling humanoid dynamics with environment geometry and contact. 
Second, a substantial portion of motions estimated from video was filtered out during curation to ensure physical reliability, reflecting the current limitations of

video-to-motion recovery; as this technology matures, future versions of PHUMA can capture a larger and more diverse range of human behaviors. \crnew{Third, our real-world comparison covers only AMASS, since the Humanoid-X-trained policy falls about three times as often in simulation (Appendix~\ref{appendix:fall_rate}), which poses a safety risk on hardware.} Finally, while PhySINK substantially reduces physical artifacts during retargeting, the motions in PHUMA do not entirely eliminate them, and further closing this gap remains an important direction for future work.

\section{Conclusion}

We introduced \datasetname, a large-scale, physically reliable humanoid locomotion dataset that overcomes the limitations of existing motion imitation pipelines. Unlike prior video-driven datasets prone to artifacts such as floating, ground penetration, and joint violations, \datasetname~combines diverse human videos with physics-aware curation and PhySINK, our \crnew{physics-aware} retargeting method, to produce motions that are both diverse and physically reliable. On the Unitree G1 and H1-2 humanoids, policies trained on \datasetname~consistently outperform those trained on LaFAN1, AMASS, and Humanoid-X, with ablations confirming that both curation and PhySINK contribute meaningfully. These advantages further extend to a real Unitree G1, where \datasetname-trained policies achieve lower tracking errors than AMASS-trained counterparts. Together, these results show that advancing humanoid motion tracking depends not just on having more data, but on having physically reliable data. PHUMA achieves both: it scales up data by incorporating diverse human videos, while ensuring physical reliability through careful curation and retargeting.

\acknowledgments{\crnew{This work was supported by the Institute for Information \& communications Technology Planning \& Evaluation (IITP) grant funded by the Korea government (MSIT) (RS-2019-II190075, Artificial Intelligence Graduate School Program (KAIST)), the National Research Foundation of Korea (NRF) grant funded by the Korea government (MSIT) (No. RS-2025-00555621), the Electronics and Telecommunications Research Institute (ETRI) grant funded by the Korean government [26CS1100, Development of Proprietary Physical AI-based Small-scale Computers and Integrated Soft Suits], and the ``Advanced GPU Utilization Support Program'' funded by the Government of the Republic of Korea (Ministry of Science and ICT).}}

\bibliography{corl2026_reference}  %

\clearpage
\appendix
\section*{\Large \textbf{Appendix}}
\makeatletter
\setlength{\@fptop}{0pt}            %
\setlength{\@fpsep}{12pt plus 4pt}  %
\setlength{\@fpbot}{0pt plus 1fil}  %
\makeatother
\setcounter{topnumber}{4}
\setcounter{totalnumber}{6}
\renewcommand{\topfraction}{0.9}
\renewcommand{\textfraction}{0.05}
\addcontentsline{toc}{section}{Appendix}

\vspace{0.4cm}
\startcontents[appendix]
\printcontents[appendix]{}{1}{\setcounter{tocdepth}{2}}
\vspace{0.3cm}

\clearpage
\section{\datasetname~Dataset}
\label{appendix:dataset}
\subsection{Data Preprocessing}
\label{appendix:preprocessing}
Before applying the retargeting procedure, it is essential to ensure that the human motion data is clean and robust, as this data serves as the target for the humanoid robot to follow. Raw motion data often contains noise from sensor errors, tracking inaccuracies, or estimation artifacts that can negatively impact retargeting. To address this, we apply the following preprocessing to filter and clean the motion data.

\subsubsection{Low-Pass Noise Filtering for Motion Data}
\label{appendix:lpf}

All motion sequences were resampled to 30 Hz. We smooth all motion channels with a zero-phase, 4th-order Butterworth low-pass filter. For root translation, the cutoff is 3 Hz; for global orientation and body pose, it is 6 Hz.

\subsubsection{Extracting Ground Contact Information}
\label{appendix:ground_contact}
\begin{figure}[h]
  \centering
  \includegraphics[width=0.6\columnwidth]{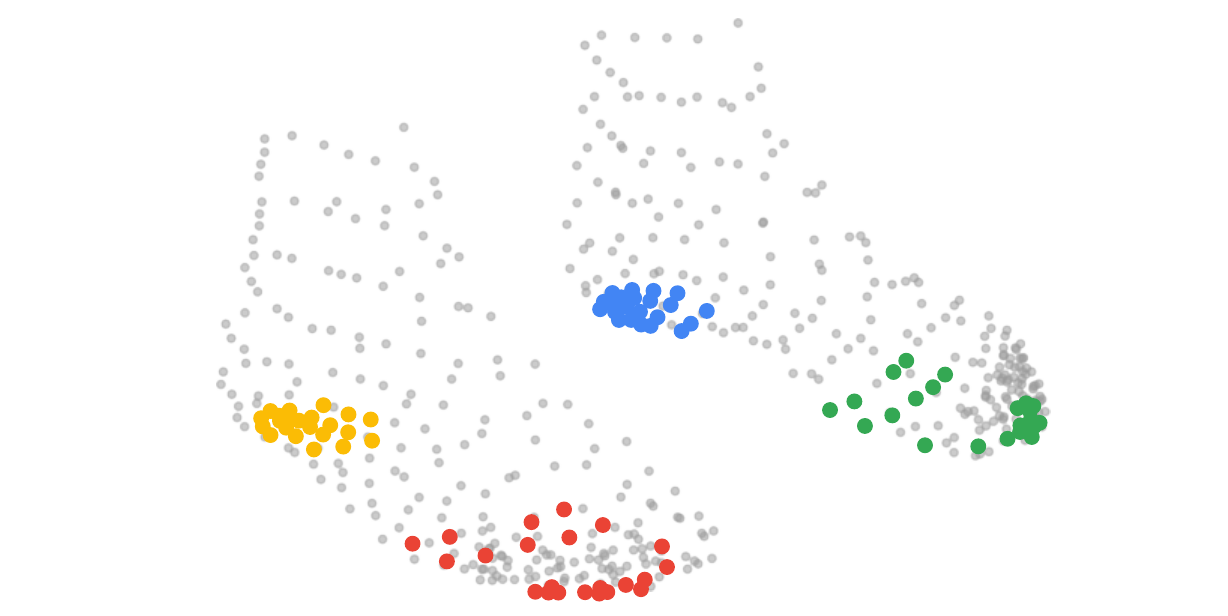}
  \caption{\textbf{SMPL-X Foot Vertices for Ground-Contact Detection}. This figure illustrates the selected foot vertices on the SMPL-X model used to detect ground contact. Blue and green points denote the left heel and left toe, while yellow and red represent the right heel and right toe, respectively. The remaining foot vertices are shown in light-gray. The clusters of colored points correspond to the specific parts of the foot that are used to check for contact with the ground, making the process more accurate and robust than using a single point.}
  \label{fig:appendix_A.1_foot_vertices}
\end{figure}

To identify foot contact, we need foot indices from a human motion source. Therefore, we identify a subset of SMPL-X foot vertices that are most indicative of ground interaction. Specifically, we select the 22 vertically lowest vertices from each foot region (left heel, left toe, right heel, right toe) in the SMPL-X default pose, totaling 88 vertices. These vertices are illustrated in Figure~\ref{fig:appendix_A.1_foot_vertices}. The vertex indices corresponding to these ground-contact points are provided in Table~\ref{tab:appendix_A.1_foot_vertices}. 

\begin{table}[h]
  \centering
  \caption{\textbf{SMPL-X foot vertex indices used for ground–contact detection.}}
  \label{tab:appendix_A.1_foot_vertices}
  \scalebox{0.85}{
      \begin{tabular}{ll}
        \toprule
        Region & Vertex indices \\ 
        \midrule
        Left heel & 8888, 8889, 8891, 8909, 8910, 8911, 8913, 8914, 8915, 8916, 8917,\\
                  & 8918, 8919, 8920, 8921, 8922, 8923, 8924, 8925, 8929, 8930, 8934 \\
        \addlinespace[2pt]
        Left toe  & 5773, 5781, 5782, 5791, 5793, 5805, 5808, 5816, 5817, 5830, 5831,\\
                  & 5859, 5860, 5906, 5907, 5908, 5909, 5912, 5914, 5915, 5916, 5917 \\
        \addlinespace[2pt]
        Right heel& 8676, 8677, 8679, 8697, 8698, 8699, 8701, 8702, 8703, 8704, 8705,\\
                  & 8706, 8707, 8708, 8709, 8710, 8711, 8712, 8713, 8714, 8715, 8716 \\
        \addlinespace[2pt]
        Right toe & 8467, 8475, 8476, 8485, 8487, 8499, 8502, 8510, 8511, 8524, 8525,\\
                  & 8553, 8554, 8600, 8601, 8602, 8603, 8606, 8608, 8609, 8610, 8611 \\
        \bottomrule
      \end{tabular}
  }
\end{table}

\begin{figure}[t]
  \centering
  \includegraphics[width=1\columnwidth]{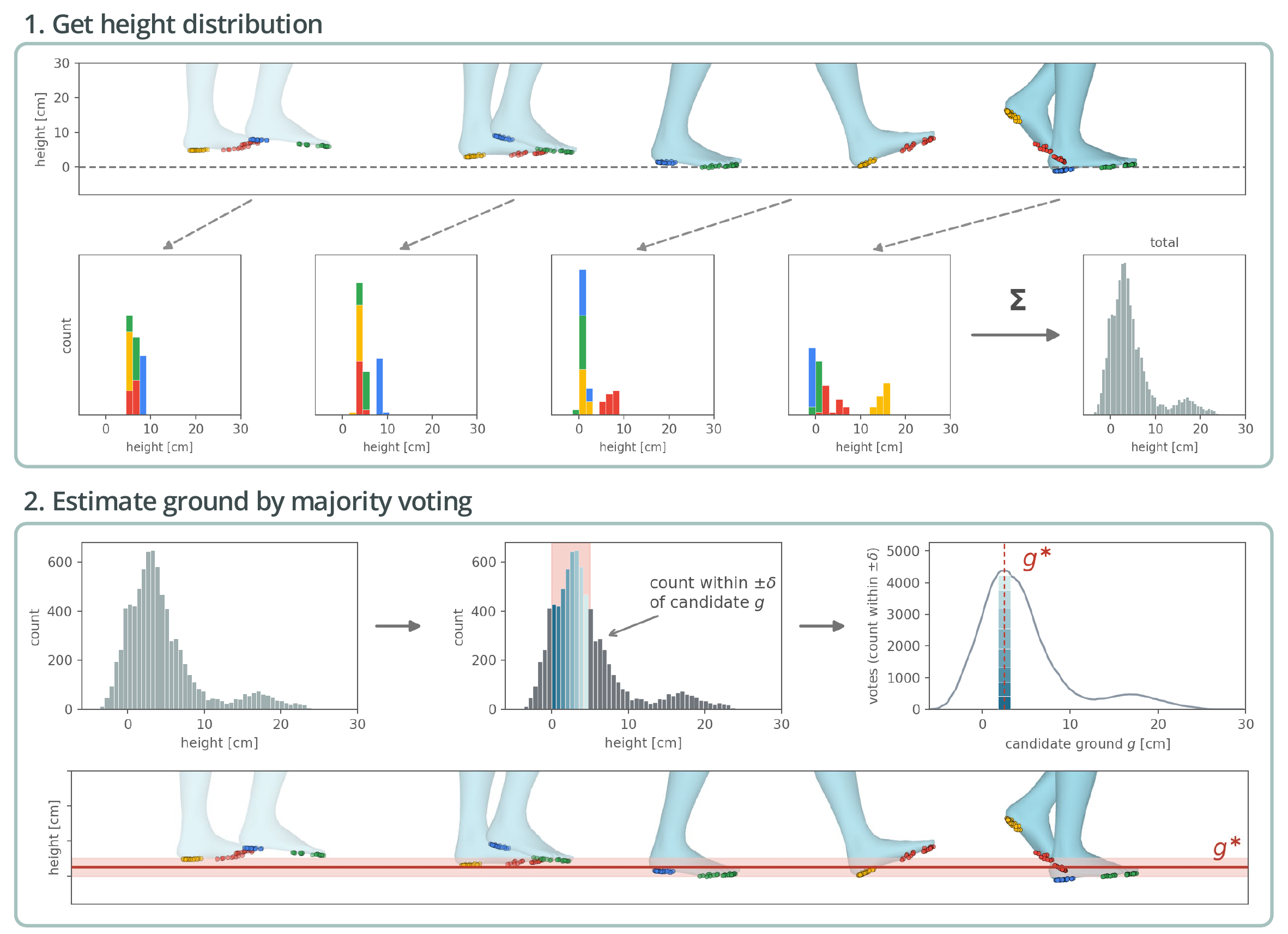}
  \caption{\textbf{Procedure of Majority-Vote Ground Estimation.} (1) shows how the foot-height distribution is obtained across all frames of a single human motion trajectory, and (2) shows how the ground height is selected from this distribution by majority vote.}
  \label{fig:majority_vote_procedure}
\end{figure}

To correctly place a motion, it is necessary to establish a single, consistent ground plane. Simple heuristics often fail: defining the ground by the lowest foot position in the sequence can cause floating, while per-frame adjustment introduces jitter and breaks down for motions where both feet leave the ground, such as jumping or running. Our method instead uses a majority vote to find the ground height that maximizes the duration of foot contact. 
Specifically, we first collect the heights of the foot vertices (Figure~\ref{fig:appendix_A.1_foot_vertices}) across all frames (Figure~\ref{fig:majority_vote_procedure}.1). Because the estimated root translation of SMPL-X is noisy, we then select the optimal ground height $g^*$ as the one with the most foot vertices falling within the band $(g-\delta, g+\delta)$ with $\delta=2.5cm$ (Figure~\ref{fig:majority_vote_procedure}.2): 
\begin{equation}
g^* = \underset{g}{\arg\max} \sum_{f=1}^{F} \sum_{v \in \mathcal{V}} \mathbf{1}\!\left[\, |h_v^f - g| < \delta \,\right],
\end{equation}
where $\mathcal{V}$ is the set of foot vertices, $h_v^f$ denotes the height of vertex $v$ at frame $f$, and $\mathbf{1}[\cdot]$ is the indicator function.
Finally, the entire motion is shifted so that this ground sits at height zero.

\subsubsection{Filtering Motion Data by Physical Information}
\label{appendix:filtering}
To remove noisy data that cannot be corrected during retargeting, we evaluate each human motion clip using the following metrics: root jerk, foot contact score, pelvis height, and pelvis/spine-1 distance to the base of support (Table~\ref{tab:appendix_A.1_filtering_criteria}). Sub-clips that fail to meet these thresholds are discarded.

\textbf{Root jerk} represents rapid changes in root acceleration, indicative of abrupt or unnatural motions. High root jerk segments are excluded to ensure smooth and physically plausible trajectories.

\begin{figure}[h]
  \centering
  \includegraphics[width=0.9\columnwidth]{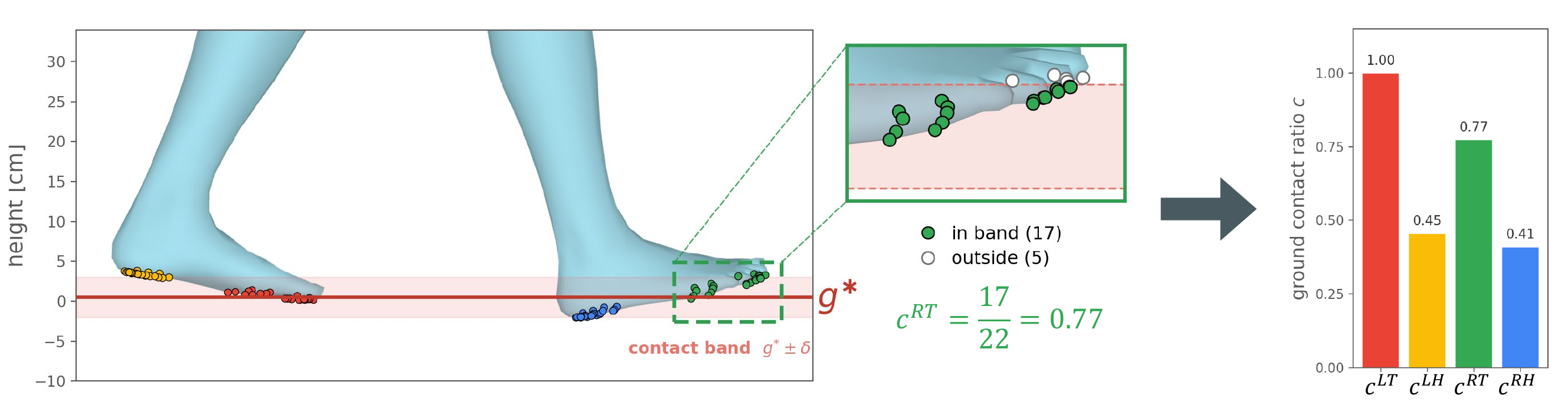}
  \caption{\textbf{Examples of Calculating the Ground-Contact Ratio.} This figure illustrates how the ground-contact ratio is computed, which is used in the foot-contact score for physics-aware curation and in the ground and skating losses for PhySINK.}
  \label{fig:foot_contact_ratio}
\end{figure}

\textbf{Foot contact score} measures the consistency and sufficiency of foot-ground interactions based on graded ground-contact signals defined by vertex proximity to the ground. Specifically, given a sub-clip with $T$ frames, the foot contact score is computed as:
\begin{equation}
    \text{Foot contact score} = \frac{1}{T} \sum_{t=1}^{T} \max\left(c^{LH}_t, c^{LT}_t, c^{RH}_t, c^{RT}_t\right),
\end{equation}
where $c^{LH}_t$, $c^{LT}_t$, $c^{RH}_t$, and $c^{RT}_t$ represent the graded ground-contact ratio at frame $t$ for the left heel, left toe, right heel, and right toe, respectively; Figure~\ref{fig:foot_contact_ratio} illustrates the ground-contact ratio in detail. A low foot contact score indicates significant penetration or floating, both of which are undesirable artifacts. Note that motions involving airborne phases, such as jumps, can easily satisfy this criterion as long as contact before and after the airborne phase is consistent.

\textbf{Pelvis height} criteria exclude segments where the humanoid is unnaturally positioned. Specifically, the minimum height criterion filters out motions that involve the humanoid being excessively crouched or lying on the ground, while the maximum height criterion eliminates segments exhibiting unnatural floating.

\textbf{Distance to the base of support} criteria ensure stable and physically plausible balance. Since the SMPL-X model's center of mass typically lies between the pelvis and spine1 joints, deviations of these joints' horizontal-plane projections from the base of support indicate imbalance or instability infeasible for humanoids. The base of support is defined as the convex hull formed by the horizontal-plane projections of the left foot, right foot, left ankle, and right ankle joints.

\begin{table}[t]
    \centering
    \caption{\textbf{Physics-aware data filtering metrics and thresholds.}}
    \label{tab:appendix_A.1_filtering_criteria}
    \begin{tabular}{ll}
        \toprule
        Metric & Threshold \\
        \midrule
        Root jerk & $< 50$ m/s$^3$ \\
        Foot contact score & $> 0.6$ \\
        Minimum pelvis height & $> 0.6$ m \\
        Maximum pelvis height & $< 1.5$ m \\
        Pelvis distance to base of support & $< 6$ cm \\
        Spine1 distance to base of support & $< 11$ cm \\
        \bottomrule
    \end{tabular}
\end{table}

\begin{crblock}
\textbf{Threshold selection.} We plot the distribution of each metric over Humanoid-X and inspect the motions along it, setting each threshold where motions stop being reproducible on flat ground (Figure~\ref{fig:curation_criteria}). The root jerk and maximum pelvis height thresholds fall at the 94th and 99th percentiles and mainly remove reconstruction artifacts, while the other four are set lower to also exclude motions outside our scope, such as sitting and off-balance motions.
\begin{figure}[h]
  \centering
  \includegraphics[width=\linewidth]{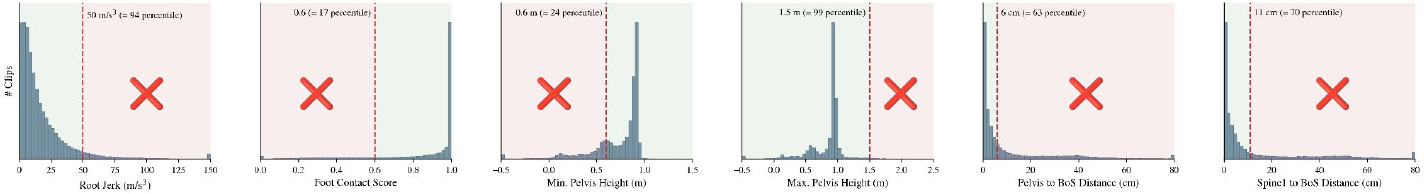}
  \caption{\textbf{Distribution of Each Curation Metric over Humanoid-X.} The dashed line marks each threshold with its percentile, and motions in the shaded region marked with a cross are discarded.}
  \label{fig:curation_criteria}
\end{figure}
\end{crblock}

\subsection{Dataset Composition and Statistics}
\label{appendix:dataset_composition}
\begin{figure}[h]
    \centering
    \begin{subfigure}[b]{0.48\textwidth}
        \includegraphics[width=\linewidth]{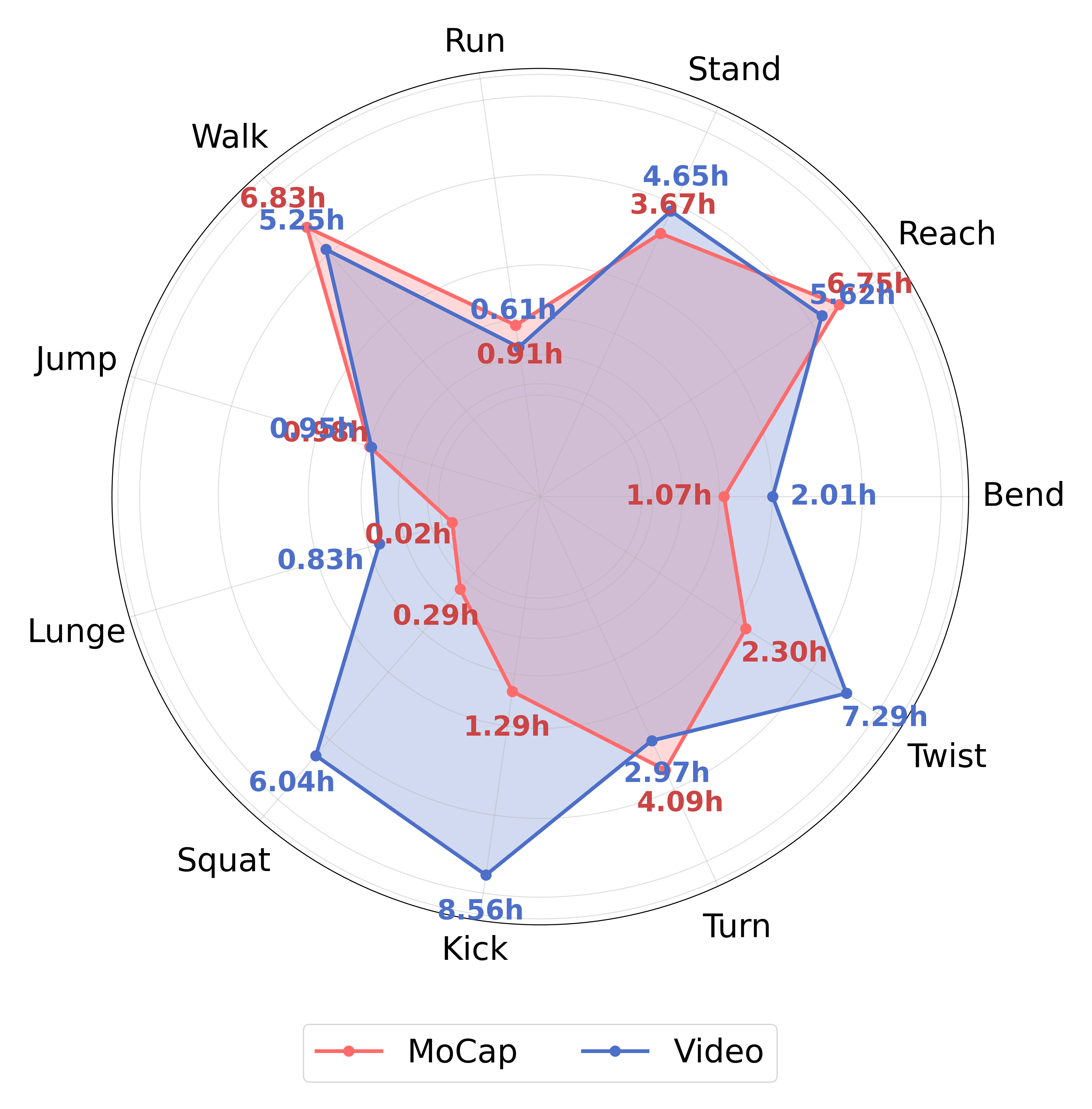}
        \caption{}
        \label{fig:mocap_vs_video}
    \end{subfigure}
    \hfill 
    \begin{subfigure}[b]{0.48\textwidth}
        \includegraphics[width=\linewidth]{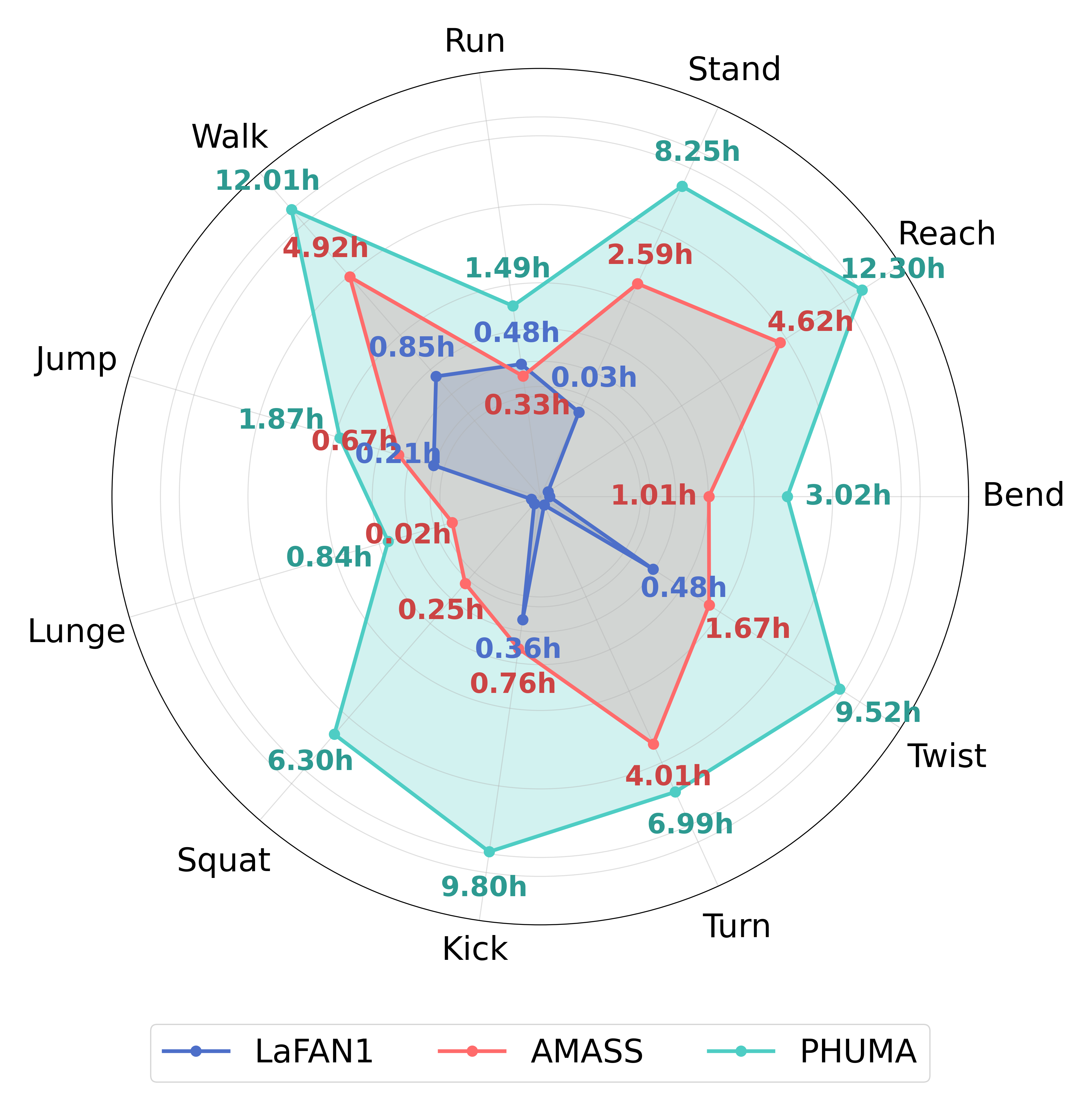}
        \caption{} 
        \label{fig:dataset_comparison}
    \end{subfigure}
    \caption{\textbf{Comparison of Motion Distributions.} (a) Comparison between MoCap and Video sources in PHUMA. (b) Comparison with other existing datasets.}
    \label{fig:appendix_motion_distribution}
\end{figure}

This section presents the detailed motion statistics of \datasetname. Because we collect motion data from diverse sources, ranging from MoCap to video, \datasetname~achieves a well-balanced motion distribution that is not dominated by any single motion type. As shown in Figure~\ref{fig:dataset_comparison}, \datasetname~exhibits substantially more balanced motion coverage than existing datasets. LaFAN1 and AMASS show uneven distributions: some motion categories are entirely absent (e.g., reach, bend, and squat), while others dominate (e.g., reach, turn, and walk). In contrast, \datasetname~provides balanced coverage across all categories, with substantially more examples per motion type. The detailed composition of \datasetname~is given in Table~\ref{tab:appendix_A.2_dataset_source}.

\begin{table}[h]
    \vspace{-4mm}
    \centering
    \caption{\textbf{Composition of the \datasetname~dataset.} A summary of the number of clips and duration for each sub-dataset, categorized by source and scene. \datasetname~aggregates these diverse sub-datasets, resulting in 73.0 hours of physically reliable motion clips.}    
    \vspace{1mm}
    \label{tab:appendix_A.2_dataset_source}
    \small
    \scalebox{0.95}{
    \begin{tabular*}{\textwidth}{l@{\extracolsep{\fill}}rrrll}
    \toprule
    Dataset & \# Clip & \# Frame & Duration & Source & Scene \\
    \midrule
    LocoMuJoCo \citep{al2023locomujoco} & 0.78K & 0.93M & 0.86h & Motion Capture & Indoor \\ %
    GRAB \citep{taheri2020grab} & 1.73K & 0.20M & 1.88h & Motion Capture & Indoor \\ %
    EgoBody \citep{zhang2022egobody} & 2.12K & 0.24M & 2.19h & Motion Capture & Indoor \\ %
    LAFAN1 \citep{harvey2020lafan} & 2.18K & 0.26M & 2.40h & Motion Capture & Indoor \\ %
    AMASS \citep{mahmood2019amass} & 21.73K & 2.25M & 20.86h & Motion Capture & Indoor \\ %
    HAA500 \citep{chung2021haa500} & 1.76K & 0.11M & 1.01h & Human Video & Outdoor \\ %
    Motion-X Video \citep{lin2023motion} & 33.04K & 3.45M & 31.98h & Human Video & Outdoor \\ %
    HuMMan \citep{cai2022humman} & 0.50K & 0.05M & 0.47h & Human Video & Indoor \\ %
    AIST \citep{tsuchida2019aist} & 1.75K & 0.18M & 1.66h & Human Video & Indoor \\ %
    IDEA400 \citep{lin2023motion} & 9.94K & 0.98M & 9.10h & Human Video & Indoor \\ %
    \textbf{\datasetname~Video} & \textbf{0.50K} & \textbf{0.06M} & \textbf{0.56h} & Human Video & Outdoor \\ %
    \midrule
    \textbf{\datasetname} & \textbf{76.01K} & \textbf{7.88M} & \textbf{72.96h} & & \\ %
    \bottomrule
    \end{tabular*}
    }
\end{table}

\section{Motion Retargeting}
\label{appendix:retargeting_details}

\crnew{This section compares PhySINK with other retargeting methods, reports its computational cost, and evaluates it as an initialization for dynamics-aware retargeting. The PhySINK loss terms are defined in Section~\ref{subsec:method_retargeting}.}

\subsection{Comparison of Retargeting Methods}
\label{appendix:qualitaive_retarget}
\begin{figure}[t]
  \centering
  \includegraphics[width=0.95\textwidth]{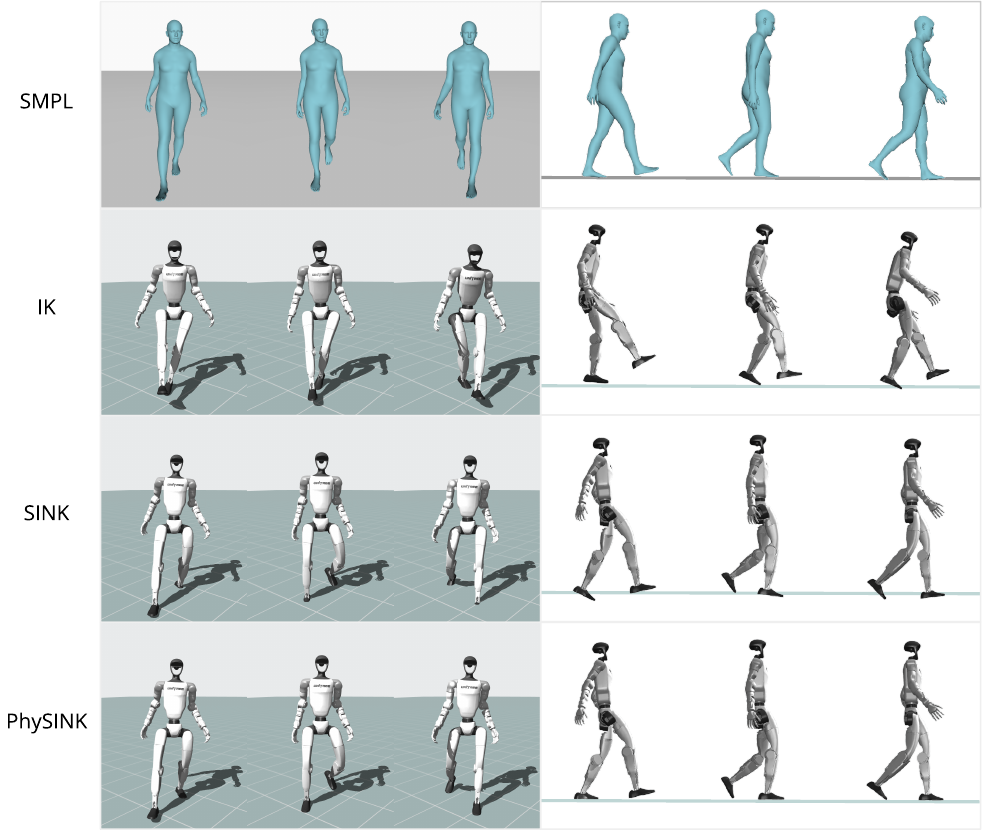}
  \caption{\textbf{Qualitative Comparison of Retargeting Methods}. This figure provides a visual comparison of human motion retargeted to a humanoid robot using the IK, SINK, and PhySINK methods. The top row shows the original human motion from the SMPL model, while the rows below show the resulting motions for each retargeting method.}
  \label{fig:appendix_B.2_retarget_method_comparison}
  \vspace{-3mm}
\end{figure}

To provide an intuitive comparison of different retargeting approaches, we present qualitative results in Figure~\ref{fig:appendix_B.2_retarget_method_comparison}. Using a walking motion as an example, we demonstrate the distinct characteristics and limitations of each method.

Traditional inverse kinematics (IK) prioritizes matching end-effector positions, such as hands and feet, from rigidly scaled human motions. However, this approach produces unnatural locomotion patterns where the humanoid appears to walk on a tightrope rather than exhibiting a natural human-like gait. This occurs because the fixed scaling cannot account for the proportional differences between human and robot morphologies.

Shape-adaptive inverse kinematics (SINK) generates more natural-looking walking motions compared to traditional IK by optimizing body proportions. However, SINK suffers from physical violations that compromise motion realism. Common issues include foot penetration through the ground surface and fixed ankle angles that result from the lack of explicit contact constraints during the retargeting process.

In contrast, our proposed PhySINK method achieves both natural movement patterns and physical reliability. The resulting motions maintain appropriate ankle angles while ensuring proper ground contact, demonstrating that PhySINK successfully balances motion naturalness with physical constraints. This improvement stems from the incorporation of explicit physical constraint terms in the optimization objective.

We further compare GMR~\citep{araujo2025gmr}, an optimization-based IK approach, with PhySINK. The two differ in how they generate the intermediate target motion. GMR scales the source motion by a heuristic: it estimates the subject's height $H_{src}$ from the first SMPL shape component ($\beta$), forms a ratio $r = H_{src} / H_{ref}$ to a reference height $H_{ref}$ (e.g., 1.8m), and applies it to manually tuned per-limb scale factors. For a limb with relative position $\mathbf{v}_{rel}$ from the pelvis and manual scale $s_{limb}$, the scaled target is $\mathbf{v}'_{target} = s_{limb} \times r \times \mathbf{v}_{rel}$, which a standard IK solver then tracks. PhySINK instead optimizes the SMPL shape parameters ($\beta$) to match the humanoid in a shared T-pose and applies them to the original motion, preserving the original joint angles while adjusting limb lengths to the robot.

\begin{figure}[h]
    \centering
    \vspace{-3mm}
    \includegraphics[width=0.85\textwidth]{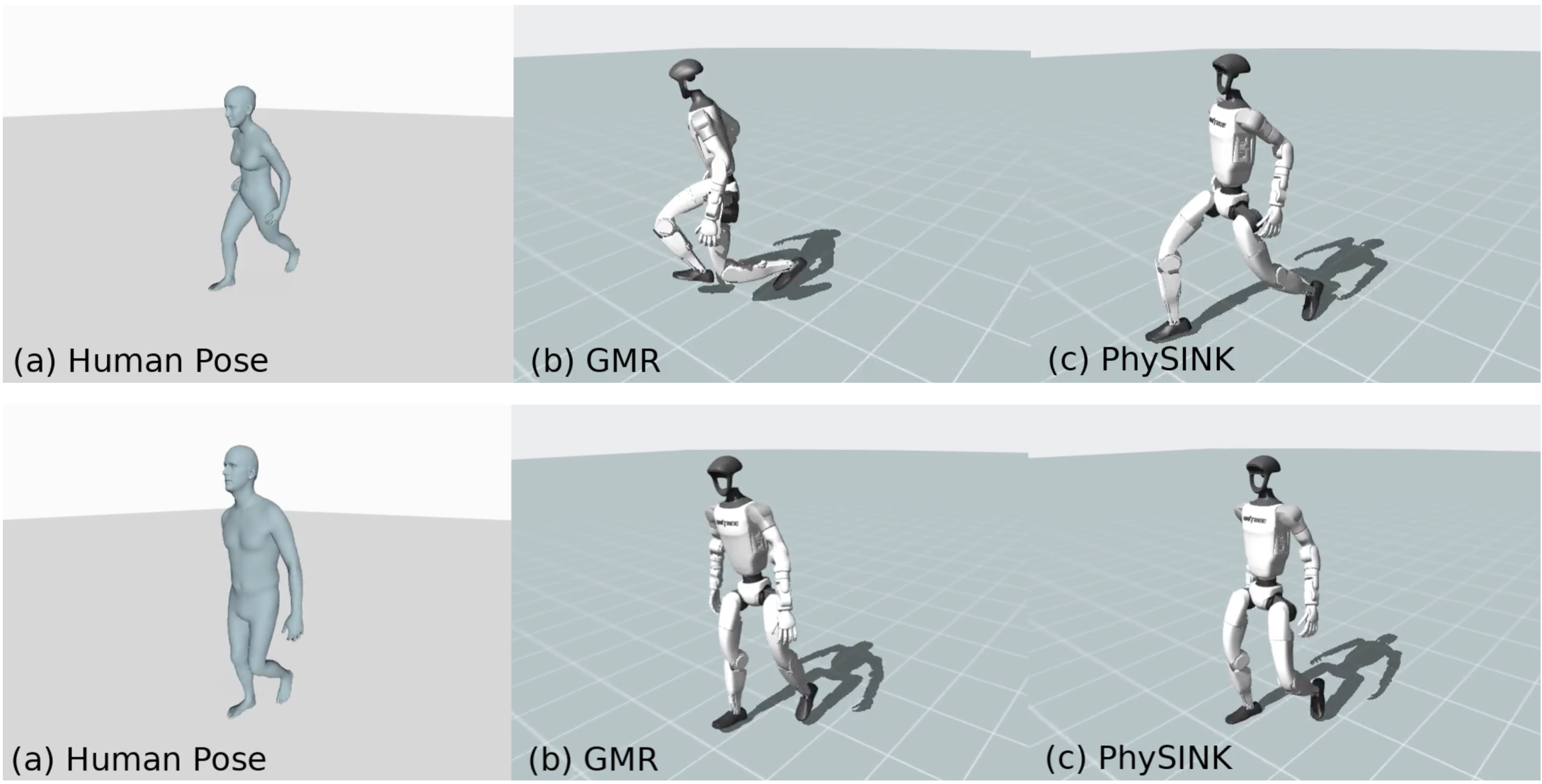}
    \caption{\textbf{Qualitative Comparison of Motion Retargeting: GMR vs. PhySINK.} Comparison showing the limitations of GMR when handling extreme human heights. The top row illustrates retargeting from a short subject, where GMR causes excessive motion compression and crouching. The bottom row illustrates retargeting from a tall subject, where GMR generates infeasible targets, resulting in artifacts like joint locking (e.g., rigid knees) and loss of ground contact. PhySINK maintains kinematic feasibility in both cases.}
    \label{fig:appendix_B.2_gmr_vs_physink}
\end{figure}

This difference matters for generalization. GMR's linear scaling works for average-sized humans but fails on diverse body shapes (Figure~\ref{fig:appendix_B.2_gmr_vs_physink}): short subjects yield undersized targets that force the robot to crouch even when standing, while tall subjects yield oversized targets beyond its reach, causing locked knees and floating feet. By applying the original joint angles with a robot-matched shape, PhySINK reproduces the intended motion regardless of the source subject's proportions.

This also affects policy training. We retarget the \datasetname~sources with GMR (excluding the pre-retargeted LaFAN1~\citep{harvey2020lafan} and LocoMuJoCo~\citep{al2023locomujoco}) and train a MaskedMimic policy under the same setup. As shown in Table~\ref{tab:appendix_B.2_imitation_performance_gmr_vs_physink}, the PhySINK-retargeted policy outperforms it on both the \datasetname~Test and Unseen Video benchmarks.

\begin{table}[h]
    \centering
    \caption{\textbf{Imitation Performance: GMR vs. PhySINK Retargeting on Unitree G1.} We evaluate the imitation performance of the MaskedMimic policy trained using datasets retargeted with GMR and PhySINK on the Unitree G1.}
    \label{tab:appendix_B.2_imitation_performance_gmr_vs_physink}
    \scalebox{0.78}{
        \begin{tabular}{l*{10}{c}}
            \toprule
            & \multicolumn{5}{c}{\datasetname \ Test} & \multicolumn{5}{c}{Unseen Video} \\
            \cmidrule(lr){2-6}
            \cmidrule(lr){7-11}
            Dataset & Total & Stationary & Angular & Vertical & Horizontal & Total & Stationary & Angular & Vertical & Horizontal \\
            \midrule
            GMR & 84.0 & 92.1 & 77.8 & 77.1 & 89.1 & 75.2 & \textbf{99.1} & 77.8 & \textbf{61.7} & 52.7 \\
            PhySINK  & \textbf{89.9} & \textbf{94.2}& \textbf{87.6}& \textbf{84.2}& \textbf{91.8}& \textbf{81.7} & 97.4& \textbf{86.7} & \textbf{61.7} & \textbf{71.4} \\
            \bottomrule
        \end{tabular}
    }
\end{table}

\begin{crblock}
\subsection{Computational Cost}
\label{appendix:compute}
Table~\ref{tab:appendix_cr_fps} reports retargeting throughput on a single RTX 3090 GPU. IK and GMR solve each frame independently, whereas SINK and PhySINK jointly optimize the entire trajectory to handle the frame-level noise of video-reconstructed motion. Since \datasetname~is constructed offline, we favor trajectory-level optimization over speed.
\begin{table}[h]
    \centering
    \caption{\textbf{Retargeting Throughput.} Frames per second of each retargeting method on a single RTX 3090 GPU.}
    \label{tab:appendix_cr_fps}
    \scalebox{0.85}{
        \begin{tabular}{lc}
            \toprule
            Retarget & FPS \\
            \midrule
            IK      & 38.4 \\
            GMR     & 25.3 \\
            SINK    & 0.95 \\
            PhySINK & 0.92 \\
            \bottomrule
        \end{tabular}
    }
\end{table}

\subsection{PhySINK as Initialization for Dynamics-Aware Retargeting}
\label{appendix:spider}
SPIDER~\citep{pan2025spider} consists of two steps: kinematic retargeting with GMR, followed by sampling-based optimization. We replace GMR with PhySINK in the first step and keep the second step unchanged. As shown in Table~\ref{tab:appendix_cr_spider}, starting from PhySINK raises the open-loop success rate on the Unseen Video motions from 4.6\% to 10.7\%.
\begin{table}[h]
    \centering
    \caption{\textbf{PhySINK as Initialization for Sampling-Based Retargeting.} Open-loop success rate (\%) on the Unseen Video motions when the sampling-based optimization of SPIDER starts from GMR, as in SPIDER, or from PhySINK.}
    \label{tab:appendix_cr_spider}
    \scalebox{0.85}{
        \begin{tabular}{lc}
            \toprule
            Retarget & Open-loop SR (\%) \\
            \midrule
            GMR $+$ Sampling (SPIDER) & 4.6 \\
            \textbf{PhySINK $+$ Sampling} & \textbf{10.7} \\
            \bottomrule
        \end{tabular}
    }
\end{table}

\end{crblock}

\section{Motion Imitation Learning}
\label{appendix:details_imitation_learning}
\subsection{Observation Space Compositions}
\label{appendix:obs_space}
This section provides detailed information about the observation space composition used in our experimental setup, as summarized in Table~\ref{tab:appendix_C.1_obs_space_maskedmimic} and Table~\ref{tab:appendix_C.1_obs_space_beyondmimic}. The observation space consists of two main components: proprioceptive states and goal states.

\subsubsection{MaskedMimic}

\textbf{Proprioceptive States.} The proprioceptive information includes root height, body positions, body rotations, body velocities, and body angular velocities. The Unitree G1 and H1-2 robots have 33 and 25 bodies, respectively. For body positions, the root body is excluded from the position measurements.

\textbf{Goal States.} The goal states comprise both relative and absolute body positions and rotations. The relative component represents the difference between the future 15 timesteps of reference motion states and the current proprioceptive state. The absolute component represents states relative to the reference motion's root position, providing a root-relative coordinate frame for the target motion.

\begin{table}[!htbp]
    \centering
    \caption{\textbf{Observation Space Dimensions of MaskedMimic.}}
    \label{tab:appendix_C.1_obs_space_maskedmimic}
    \begin{tabular}{lcc}
        \toprule
        & \multicolumn{2}{c}{Dimension} \\
        \cmidrule(lr){2-3}
        State & G1 & H1-2 \\
        \midrule
        \rowcolor{header}\multicolumn{3}{l}{(a) Proprioceptive State} \\
        \midrule
        Root height & 1 & 1 \\
        Body position & $32 \times 3$ & $24 \times 3$ \\
        Body rotation & $33 \times 6$ & $25 \times 6$ \\
        Body velocity & $33 \times 3$ & $25 \times 3$ \\
        Body angular velocity & $33 \times 3$ & $25 \times 3$ \\
        \midrule
        \rowcolor{header}\multicolumn{3}{l}{(b) Goal State} \\
        \midrule
        Relative body position & $33 \times 15 \times 3$ & $25 \times 15 \times 3$ \\
        Absolute body position & $33 \times 15 \times 3$ & $25 \times 15 \times 3$ \\
        Relative body rotation & $33 \times 15 \times 6$ & $25 \times 15 \times 6$ \\
        Absolute body rotation & $33 \times 15 \times 6$ & $25 \times 15 \times 6$ \\
        Time & $33 \times 15 \times 1$ & $25 \times 15 \times 1$ \\
        \midrule
        \textbf{Total dim} & 9898 & 7498 \\
        \bottomrule
    \end{tabular}
\end{table}

\subsubsection{BeyondMimic}
\label{appendix:obs_space_beyondmimic}
  \textbf{Proprioceptive States.} The proprioceptive information includes the previous action, the base angular velocity in the body frame, and the DoF position and DoF velocity vectors.

  \textbf{Goal States.} The goal states comprise an instantaneous component and a future component. The instantaneous component provides the reference motion's DoF position, DoF velocity, and 6D torso orientation. 
  
  Unlike the original BeyondMimic, we augment the goal state with the reference motion's future trajectory. At each future frame, we add the components listed in Table~\ref{tab:appendix_C.1_obs_space_beyondmimic}. We sample $K=20$ such frames, linearly spaced over a horizon of $H=95$ control steps ($1.9\,\text{s}$ at $50\,\text{Hz}$), forming a $20 \times 78$ tensor. A separate motion encoder then projects this tensor to a $128$-D embedding, which is concatenated with the rest of the observation. Including this future component yields a substantially higher success rate than relying on the instantaneous goal alone, in both IsaacSim and MuJoCo and on both the PHUMA Test and Unseen Video splits (Table~\ref{tab:appendix_C.1_beyondmimic_eval}).

\begin{table}[!htbp]
      \centering
      \caption{\textbf{Observation Space Dimensions of BeyondMimic.}}
      \label{tab:appendix_C.1_obs_space_beyondmimic}
      \begin{tabular}{lc}
          \toprule
          State & Dimension (G1) \\
          \midrule
          \rowcolor{header}\multicolumn{2}{l}{(a) Proprioceptive State} \\
          \midrule
          Previous action & 29 \\
          Base angular velocity & 3 \\
          DoF position & 29 \\
          DoF velocity & 29 \\
          \midrule
          \rowcolor{header}\multicolumn{2}{l}{(b) Goal State} \\
          \midrule
          \rowcolor{subheader}\multicolumn{2}{l}{(b.1) Instantaneous} \\
          DoF position & 29 \\
          DoF velocity & 29 \\
          Torso rotation & 6 \\
          \midrule
          \rowcolor{subheader}\multicolumn{2}{l}{(b.2) Future Motion Targets (motion-encoder input)} \\
          Root height & $20 \times 1$ \\
          Root roll/pitch & $20 \times 2$ \\
          Base linear velocity & $20 \times 3$ \\
          Base yaw rate & $20 \times 1$ \\
          DoF position & $20 \times 29$ \\
          Local key-body position & $20 \times 14 \times 3$ \\
          \midrule
          \textbf{Total dim} & 282 \\
          \bottomrule
      \end{tabular}
  \end{table}
   
\begin{table}[!htbp]
      \centering
      \caption{\textbf{Sim-to-Sim Results of PHUMA-trained BeyondMimic Policies.} This table demonstrates the effect of the future-trajectory observation on the policy for the Unitree G1, evaluated on the PHUMA Test and Unseen Video splits. SR is the whole-body success rate at $0.5\,\text{m}$ (\%); $E_{\text{g-mpjpe}}$, $E_{\text{l-mpjpe}}$, $E_{\text{acc}}$, and $E_{\text{vel}}$ are tracking errors (mm), following the convention of~\citep{luo2023phc, he2024h2o, he2025asap} and averaged over successful motions.}
      \label{tab:appendix_C.1_beyondmimic_eval}
      \scalebox{0.85}{
          \begin{tabular}{ccccccccccc}
            \toprule
            \multirow{2}{*}{Future}
              & \multicolumn{5}{c}{IsaacSim}
              & \multicolumn{5}{c}{MuJoCo} \\
            \cmidrule(lr){2-6} \cmidrule(lr){7-11}
              & SR $\uparrow$ & $E_{\text{g-mpjpe}}\downarrow$ & $E_{\text{l-mpjpe}}\downarrow$ & $E_{\text{acc}}\downarrow$ &
  $E_{\text{vel}}\downarrow$
              & SR $\uparrow$ & $E_{\text{g-mpjpe}}\downarrow$ & $E_{\text{l-mpjpe}}\downarrow$ & $E_{\text{acc}}\downarrow$ &
  $E_{\text{vel}}\downarrow$ \\
            \midrule
            \rowcolor{header}\multicolumn{11}{l}{(a) PHUMA test} \\
            \midrule
            \xmark & 47.15          & 115.78          & 32.70          & \textbf{1.15} & 4.29
                   & 46.91          & 156.64          & 32.92          & \textbf{1.22} & 4.36 \\
            \cmark & \textbf{78.77} & \textbf{112.00} & \textbf{28.28} & 1.18          & \textbf{4.14}
                   & \textbf{76.02} & \textbf{142.79} & \textbf{28.98} & 1.24          & \textbf{4.18} \\
            \midrule
            \rowcolor{header}\multicolumn{11}{l}{(b) Unseen Video} \\
            \midrule
            \xmark & 47.82          & 123.71          & 31.93          & \textbf{1.07} & 4.93
                   & 47.42          & 122.89          & 30.60          & \textbf{1.09} & 4.85 \\
            \cmark & \textbf{71.63} & \textbf{121.69} & \textbf{24.87} & 1.11          & \textbf{4.78}
                   & \textbf{70.44} & \textbf{122.68} & \textbf{23.79} & 1.14          & \textbf{4.73} \\
            \bottomrule
          \end{tabular}
      }
  \end{table}

\subsection{Reward Function}
  \label{appendix:reward_function}
  The reward function used for training the tracking policy consists of multiple components, as detailed in Table~\ref{tab:appendix_reward_terms} and Table~\ref{tab:appendix_reward_terms_beyondmimic}. The overall reward structure comprises
   two main categories: motion tracking task rewards and regularization rewards.

  \subsubsection{MaskedMimic}

  \textbf{Motion Tracking Rewards.} These components encourage the policy to match the reference motion by providing higher rewards when the robot's proprioceptive states closely resemble the target motion states.

  \textbf{Regularization Rewards.} To promote smooth and stable motion execution, we include regularization terms that penalize undesirable behaviors. Specifically, we augment the standard MaskedMimic reward formulation with action rate
  penalties that discourage large changes between consecutive actions, helping to ensure smooth joint movements and prevent abrupt motion transitions.

  \begin{table}[!htbp]
      \centering
      \caption{\textbf{Reward Function Terms of MaskedMimic.}}
      \label{tab:appendix_reward_terms}
      \begin{tabular}{lll}
          \toprule
          Term & Expression & Weight \\
          \midrule
          \rowcolor{gray!15}\multicolumn{3}{l}{(a) Task} \\
          \midrule
          Global body position & $\exp(-100 \cdot \lVert p_t - \hat{p}_t \rVert^2_2)$ & 0.5 \\
          Root height & $\exp(-100 \cdot (h^{\text{root}}_t - \hat{h}^{\text{root}}_t)^2)$ & 0.2 \\
          Global body rotation & $\exp(-10 \cdot \lVert \theta_t \ominus \hat{\theta}_t \rVert^2_2)$ & 0.3 \\
          Global body velocity & $\exp(-0.5 \cdot \lVert v_t - \hat{v}_t \rVert^2_2)$ & 0.1 \\
          Global body angular velocity & $\exp(-0.1 \cdot \lVert \omega_t - \hat{\omega}_t \rVert^2_2)$ & 0.1 \\
          \midrule
          \rowcolor{gray!15}\multicolumn{3}{l}{(b) Regularization} \\
          \midrule
          Power consumption & $\lVert F \odot \dot{q} \rVert_1$ & -1e-05 \\
          Action rate & $\lVert a_t - a_{t-1} \rVert^2_2$ & -0.2 \\
          \bottomrule
      \end{tabular}
  \end{table}

  \subsubsection{BeyondMimic}
  \label{appendix:reward_function_beyondmimic}

  \textbf{Motion Tracking Rewards.} BeyondMimic decomposes motion tracking into two components: tracking the global reference motion (position, rotation, and linear/angular velocity), and tracking the relative reference motion, which factors out the root's global horizontal translation (and heading) and re-anchors the body targets to the robot's own root.

  \textbf{Regularization Rewards.} To promote smooth and stable motion execution we use the same regularization terms used in BeyondMimic as in Table \ref{tab:appendix_reward_terms_beyondmimic}.

  \begin{table}[!htbp]
      \centering
      \caption{\textbf{Reward Function Terms of BeyondMimic.}}
      \label{tab:appendix_reward_terms_beyondmimic}
      \begin{tabular}{lll}
          \toprule
          Term & Expression & Weight \\
          \midrule
          \rowcolor{gray!15}\multicolumn{3}{l}{(a) Motion Tracking} \\
          \midrule
          Reference body position & $\exp(-\lVert p^{\text{ref}}_t - \hat{p}^{\text{ref}}_t \rVert^2_2 \,/\, 0.3^2)$ & 0.5 \\
          Reference body rotation & $\exp(-\lVert \theta^{\text{ref}}_t \ominus \hat{\theta}^{\text{ref}}_t \rVert^2_2 \,/\, 0.4^2)$ & 0.5 \\
          Global body linear velocity & $\exp\!\left(-\tfrac{1}{B}\sum_{b} \lVert v^{b}_t - \hat{v}^{b}_t \rVert^2_2 \,/\, 1.0^2\right)$ & 1.0 \\
          Global body angular velocity & $\exp\!\left(-\tfrac{1}{B}\sum_{b} \lVert \omega^{b}_t - \hat{\omega}^{b}_t \rVert^2_2 \,/\, \pi^2\right)$ & 1.0 \\
          Relative body position & $\exp\!\left(-\tfrac{1}{B}\sum_{b} \lVert p^{b}_t - \hat{p}^{b}_t \rVert^2_2 \,/\, 0.3^2\right)$ & 1.0 \\
          Relative body rotation & $\exp\!\left(-\tfrac{1}{B}\sum_{b} \lVert \theta^{b}_t \ominus \hat{\theta}^{b}_t \rVert^2_2 \,/\, 0.4^2\right)$ & 1.0 \\
          \midrule
          \rowcolor{gray!15}\multicolumn{3}{l}{(b) Regularization} \\
          \midrule
          Action rate & $\lVert a_t - a_{t-1} \rVert^2_2$ & -0.1 \\
          DoF position-limit violation & $\sum_j \big(\,\mathrm{ReLU}(q^j_t - q^j_{\max,0.9}) + \mathrm{ReLU}(q^j_{\min,0.9} - q^j_t)\,\big)$ & -100 \\
          Undesired contacts & $\sum_b \mathbf{1}\!\left[\,\lVert F^b_t \rVert_2 > 1\,\text{N}\,\right]$ & -0.5 \\
          \bottomrule
      \end{tabular}
  \end{table}

\subsection{PPO Hyperparameters}
  \label{appendix:ppo}
  The detailed hyperparameter configurations used for PPO training in MaskedMimic and BeyondMimic are provided in Table~\ref{tab:appendix_ppo_hyperparameters} and Table~\ref{tab:appendix_ppo_hyperparameters_beyondmimic}, respectively.

  \begin{table}[!htbp]
    \centering
    \caption{\textbf{PPO Hyperparameter Values for Model Training of MaskedMimic.}}
    \label{tab:appendix_ppo_hyperparameters}
    \begin{tabular}{lc}
    \toprule
    \textbf{Hyperparameter} & \textbf{Value} \\
    \midrule
    Optimizer & Adam \\
    Num envs & 8192 \\
    Mini Batches & 32 \\
    Learning epochs & 1 \\
    Entropy coefficient & 0.0 \\
    Value loss coefficient & 0.5 \\
    Clip param & 0.2 \\
    Max grad norm & 50.0 \\
    Init noise std & -2.9 \\
    Actor learning rate & 2e-5 \\
    Critic learning rate & 1e-4 \\
    GAE decay factor($\lambda$) & 0.95 \\
    GAE discount factor($\gamma$) & 0.99 \\
    \midrule
    Actor Transformer dimension & 512 \\
    Actor layers & 4 \\
    Actor heads & 4 \\
    Critic MLP size & [1024, 1024, 1024, 1024] \\
    Activation & ReLU \\
    \bottomrule
    \end{tabular}
\end{table}

   \begin{table}[!htbp]
      \centering
      \caption{\textbf{PPO Hyperparameter Values for Model Training of BeyondMimic.}}
      \label{tab:appendix_ppo_hyperparameters_beyondmimic}
      \begin{tabular}{lc}
      \toprule
      \textbf{Hyperparameter} & \textbf{Value} \\
      \midrule
      Optimizer & AdamW \\
      Num envs & 8192 \\
      Steps per env & 24 \\
      Mini batches & 4 \\
      Learning epochs & 8 \\
      Entropy coefficient & 0.005 \\
      Value loss coefficient & 1.0 \\
      Clip param & 0.2 \\
      Max grad norm & 1.0 \\
      Init noise std & 1.0 \\
      Actor learning rate & 1e-5 \\
      Critic learning rate & 1e-5 \\
      GAE decay factor($\lambda$) & 0.95 \\
      GAE discount factor($\gamma$) & 0.99 \\
      \midrule
      Actor MLP size & [768, 512, 256] \\
      Critic MLP size & [768, 512, 256] \\
      Motion encoder hidden dim & 60 \\
      Motion encoder output dim & 128 \\
      Activation & SiLU \\
      \bottomrule
      \end{tabular}
  \end{table}

\subsection{Domain Randomization}
  \label{appendix:domain_rand}
  To bridge the simulation-to-reality gap and improve robustness to model and sensor inaccuracies, we apply domain randomization during training. The complete set of randomized parameters is summarized in
  Table~\ref{tab:appendix_domain_rand_beyondmimic}.

  \subsubsection{BeyondMimic}
  \label{appendix:domain_rand_beyondmimic}

  We adopt the domain randomization scheme of BeyondMimic~\cite{truong2025beyondmimic}, covering actuator gains, base inertial properties, contact properties, external pushes, initial-pose perturbations, and observation noise. Noise is added only to
  the actor observation stream; the critic stream is noise-free. All randomization and noise are disabled at evaluation time. The exact ranges and schedules are listed in Table~\ref{tab:appendix_domain_rand_beyondmimic}.

  \begin{table}[!htbp]
      \centering
      \caption{\textbf{Domain Randomization Ranges of BeyondMimic.}}
      \label{tab:appendix_domain_rand_beyondmimic}
      \begin{tabular}{lc}
      \toprule 
      \textbf{Parameter} & \textbf{Range} \\
      \midrule
      P-gain scale $s_P$         & $\mathcal{U}(0.9, 1.1)$ \\
      D-gain scale $s_D$         & $\mathcal{U}(0.9, 1.1)$ \\
      DoF position bias          & $\mathcal{U}(-0.025, 0.025)\,\text{rad}$ \\
      \midrule
      Base CoM offset $x$        & $\mathcal{U}(-0.025, 0.025)\,\text{m}$ \\
      Base CoM offset $y, z$     & $\mathcal{U}(-0.05, 0.05)\,\text{m}$ \\
      Static friction            & $\mathcal{U}(0.3, 1.6)$ \\
      Dynamic friction           & $\mathcal{U}(0.3, 1.2)$ \\
      Restitution                & $\mathcal{U}(0.0, 0.5)$ \\
      \midrule 
      Push interval              & $\mathcal{U}(1.0, 3.0)\,\text{s}$ \\
      Push linear velocity       & $\mathcal{U}(\pm 0.5, \pm 0.5, \pm 0.2)\,\text{m/s}$ \\
      Push angular velocity      & $\mathcal{U}(\pm 0.52, \pm 0.52, \pm 0.78)\,\text{rad/s}$ \\
      \midrule
      Initial DoF position       & $\mathcal{U}(-0.1, 0.1)\,\text{rad}$ \\
      Initial root position      & $\mathcal{U}(\pm 0.05, \pm 0.05, \pm 0.01)\,\text{m}$ \\
      Initial root rotation      & $\mathcal{U}(\pm 0.1, \pm 0.1, \pm 0.2)\,\text{rad}$ \\
      Initial root linear vel.   & $\mathcal{U}(\pm 0.1, \pm 0.1, \pm 0.05)\,\text{m/s}$ \\
      Initial root angular vel.  & $\mathcal{U}(\pm 0.1, \pm 0.1, \pm 0.1)\,\text{rad/s}$ \\
      \midrule
      Obs.\ noise: base angular vel.\ ($\sigma$)        & $0.2\,\text{rad/s}$ \\
      Obs.\ noise: DoF position ($\sigma$)              & $0.01\,\text{rad}$ \\
      Obs.\ noise: DoF velocity ($\sigma$)              & $0.5\,\text{rad/s}$ \\
      Obs.\ noise: motion ref orientation ($\sigma$)    & $0.05$ \\
      \bottomrule
      \end{tabular}
  \end{table}

\section{Experiment Details}
\label{appendix:experiment_details}
\subsection{Self-Collected Video Dataset}
\label{appendix:self_collected_video}
To ensure fair evaluation of imitation performance on unseen motions, we create a custom evaluation dataset using self-collected video recordings. This dataset contains motions uniformly distributed across the 11 motion types shown in Figure~\ref{fig:dataset_comparison}, providing balanced coverage for comprehensive performance assessment.

The dataset creation process follows three main steps: (1) recording videos of human performers executing each motion type, (2) converting videos into SMPL human motion parameters using a video-to-motion model, and (3) retargeting the human motions to humanoid robot motions using our PhySINK method.

\begin{figure}[htbp]
    \centering
    \includegraphics[width=\textwidth]{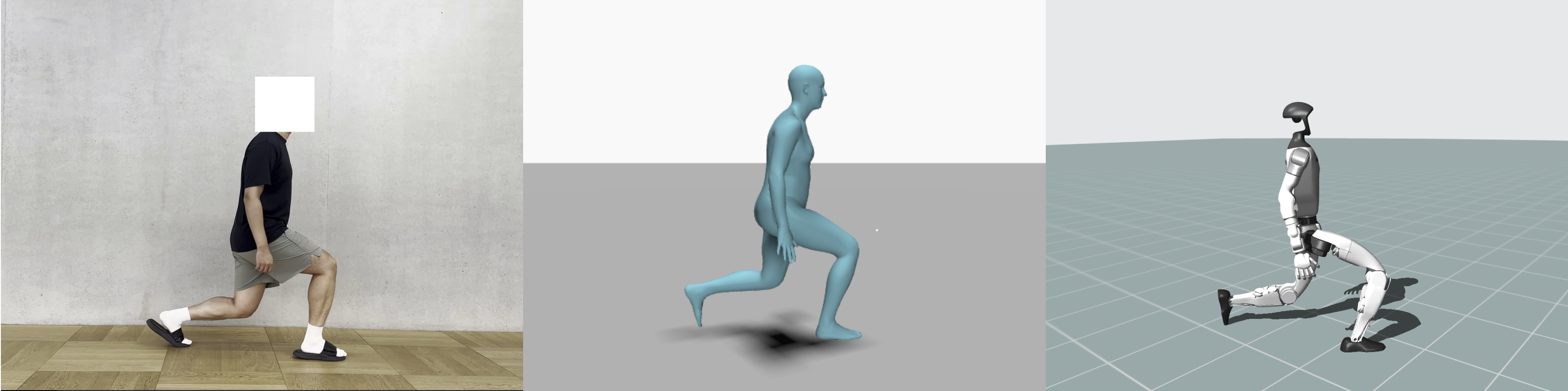}
    \caption{\textbf{Overview of the Self-collected Data Pipeline}. This figure illustrates the three main steps of our data collection pipeline: (left) a self-recorded video of a human motion, (center) the motion extracted using a video-to-motion model, and (right) the final motion retargeted to a humanoid robot.}
    \label{fig:appendix_D.1_data_collection_video}
\end{figure}

First, we record videos covering all 11 motion categories, collecting a uniform distribution for each type. We then apply the TRAM video-to-motion model~\citep{wang2024tram} to extract SMPL motion parameters from the recorded videos. Finally, we process these SMPL motions with PhySINK retargeting to generate physically reliable humanoid motions. Example results from this dataset are illustrated in Figure~\ref{fig:appendix_D.1_data_collection_video}.

This self-collected evaluation set ensures that our performance assessments are conducted on completely unseen motions that were not influenced by any training data sources. \crnew{Appendix~\ref{appendix:gmr_reference} repeats the main comparisons on evaluation motions retargeted with GMR.}

\subsection{Success Rate Threshold Analysis}
\label{appendix:performance_threshold}
To demonstrate the limitations of the conventional success rate threshold, we evaluate imitation performance using both the standard 0.5m threshold and our proposed stricter 0.15m threshold. This comparison reveals the true quality differences between policies trained on different datasets.

Tables~\ref{tab:appendix_D.2_tracking_comparison_on_dataset_PHUMA_test} and~\ref{tab:appendix_D.2_tracking_comparison_on_dataset_unseen_video} present the results for both threshold settings. Under the loose 0.5m threshold, policies trained on different datasets show relatively similar success rates, with differences appearing modest. However, when evaluated with the stricter 0.15m threshold, performance differences become substantially more pronounced.

These results confirm that \datasetname-trained policies achieve more precise motion tracking, producing imitations that remain accurate even under stringent evaluation criteria. The threshold analysis validates our choice to adopt the 0.15m threshold as a more meaningful measure of imitation quality.
\begin{table}[h]
    \centering
    \caption{\textbf{Performance Comparison based on Success Threshold in \datasetname~ Test.}}
    \label{tab:appendix_D.2_tracking_comparison_on_dataset_PHUMA_test}
    \scalebox{0.7}{
        \begin{tabular}{l*{11}{c}}
            \toprule
            & & \multicolumn{5}{c}{Success Threshold=0.15m} & \multicolumn{5}{c}{Success Threshold=0.5m} \\
            \cmidrule(lr){3-7}
            \cmidrule(lr){8-12}
            Dataset & Hours & Total & Stationary & Angular & Vertical & Horizontal & Total & Stationary & Angular & Vertical & Horizontal  \\
            \midrule
            \rowcolor{header}\multicolumn{12}{l}{\textbf{(a) G1}} \\
            \midrule
            LaFAN1                & 2.4   & 46.1 & 66.1& 36.2 & 24.0 & 42.5 & 74.8 &87.8& 69.2 & 47.1 & 72.6 \\
            AMASS                 &20.9 & 76.2 & 88.5 & 72.1 & 56.8 & 66.8 & 90.2 &95.0 & 87.9 & 81.1 & 83.7 \\
            Humanoid-X            & 231.4 & 50.6 & 78.4 & 43.0 & 26.0 & 31.8 & 78.4 & 91.3 & 72.9 & 59.5 & 65.9 \\
            \textbf{\datasetname} & 73.0  & \textbf{92.7} & \textbf{95.6} & \textbf{91.7} & \textbf{86.0} & \textbf{85.6} & \textbf{97.1} & \textbf{98.7} &  \textbf{96.5} & \textbf{94.4} & \textbf{92.5} \\
            \midrule
            \rowcolor{header} \multicolumn{12}{l}{\textbf{(b) H1-2}} \\
            \midrule
            LaFAN1                & 2.4   & 62.0 & 79.3 & 54.7 & 26.6 & 58.9 & 70.8 & 92.4 & 66.7 & 56.4 & 68.2 \\
            AMASS                 &20.9 &54.4& 74.9 & 45.9 & 17.2 & 49.6 & 70.4& 86.3 & 62.6 & 41.4 & 65.9 \\
            Humanoid-X            & 231.4 & 49.7 & 74.6 & 40.4 & 17.0 & 37.3 & 54.8 & 78.5 & 45.2 & 22.1 & 43.2 \\
            \textbf{\datasetname} & 73.0  & \textbf{82.7} & \textbf{91.5} & \textbf{79.5} & \textbf{68.1} & \textbf{68.4}& \textbf{92.0}& \textbf{96.6} &  \textbf{89.7} & \textbf{85.6} & \textbf{79.4} \\
            \bottomrule
        \end{tabular}
    }
\end{table}

\begin{table}[h]
    \centering
    \caption{\textbf{Performance Comparison based on Success Threshold in Unseen Video.}}
    \label{tab:appendix_D.2_tracking_comparison_on_dataset_unseen_video}
    \scalebox{0.7}{
        \begin{tabular}{l*{11}{c}}
            \toprule
            & & \multicolumn{5}{c}{Success Threshold=0.15m} & \multicolumn{5}{c}{Success Threshold=0.5m} \\
            \cmidrule(lr){3-7}
            \cmidrule(lr){8-12}
            Dataset & Hours & Total & Stationary & Angular & Vertical & Horizontal & Total & Stationary & Angular & Vertical & Horizontal  \\
            \midrule
            \rowcolor{header}\multicolumn{12}{l}{\textbf{(a) G1}} \\
            \midrule
            LaFAN1                & 2.4   & 28.4 & 46.9 & 28.4 & 19.6 & 10.5 & 78.2 & 85.5 & 70.8 & 76.3 & 80.8 \\
            AMASS                 &20.9 & 70.2 & 90.7 & 75.0 & 62.7 & 44.1 & 92.3 &99.2 & 92.1 & 82.1 & 88.0 \\
            Humanoid-X            & 231.4 & 39.1 & 78.0 & 39.6 & 23.0 & 6.5 & 84.1 & 98.3 & 79.9 & 76.0 & 76.2 \\
            \textbf{\datasetname} & 73.0  & \textbf{82.9} & \textbf{96.7} &  \textbf{88.0} & \textbf{71.8} & \textbf{67.1} & \textbf{93.7} & \textbf{100.0} &  \textbf{96.8} & \textbf{85.9} & \textbf{84.7} \\
            \midrule
            \rowcolor{header} \multicolumn{12}{l}{\textbf{(b) H1-2}} \\
            \midrule
            LaFAN1                & 2.4   & 70.8 & 92.4 & 66.7 & 56.4 & 68.2 & 85.5 & 97.5 & 79.0 & 77.5 & 90.0 \\
            AMASS                 &20.9 &64.3& 87.3 & 59.7 & 46.0 & 63.9 & 80.4& 93.3 & 69.9 & 72.8 & 89.0 \\
            Humanoid-X            & 231.4 & 60.5 & 88.3 & 60.0 & 48.7 & 39.7 & 68.7 & 93.3 & 65.1 & 60.2 & 50.5 \\
            \textbf{\datasetname} & 73.0  & \textbf{78.6}& \textbf{97.5} &  \textbf{76.8} & \textbf{74.5} & \textbf{63.8}& \textbf{89.9}& \textbf{99.2} &  \textbf{89.4} & \textbf{84.6} & \textbf{83.9} \\
            \bottomrule
        \end{tabular}
    }
\end{table}

\subsection{Cross-Simulator Generalization Analysis}
\label{appendix:sim_to_sim}
To evaluate whether \datasetname~benefits cross-simulator transfer, we also trained the KungfuBot~\citep{han2025kungfubot2} policy in Isaac Gym. We compared two training configurations: one using AMASS with SINK retargeting and another using \datasetname. For both configurations, we followed the standard KungfuBot~\citep{han2025kungfubot2} teacher-student training pipeline and evaluation metrics.
Tables~\ref{tab:appendix_D.3_kungfubot_teacher} and~\ref{tab:appendix_D.3_kungfubot_student} present the teacher and student policy performance in the training simulator (Isaac Gym). The results show that policies trained with \datasetname~achieve superior imitation performance, demonstrating that \datasetname's benefits extend to other motion imitation algorithms.
Importantly, this performance advantage also transfers to the evaluation simulator. Table~\ref{tab:appendix_D.3_sim-to-sim} shows the zero-shot performance in MuJoCo, where the \datasetname-trained policy maintains its superiority across motion categories, indicating robust cross-simulator generalization.
\begin{table}[h]
    \centering
    \caption{\textbf{Imitation Performance of the KungfuBot Teacher Policy in Isaac Gym.} We evaluate the imitation performance of KungfuBot teacher policy with unseen motions on the Unitree G1.}
    \label{tab:appendix_D.3_kungfubot_teacher}
    \scalebox{0.78}{
        \begin{tabular}{l*{10}{c}}
            \toprule
            & \multicolumn{5}{c}{\datasetname \ Test} & \multicolumn{5}{c}{Unseen Video} \\
            \cmidrule(lr){2-6}
            \cmidrule(lr){7-11}
            Dataset & Total & Stationary & Angular & Vertical & Horizontal & Total & Stationary & Angular & Vertical & Horizontal \\
            \midrule
            AMASS & 77.8 & 90.8 & 71.2 & 58.3 & 84.7 & 83.5 & \textbf{100.0} & 83.3 & 74.5 & 72.5 \\
            \datasetname  & \textbf{91.0} & \textbf{96.4}& \textbf{88.3}& \textbf{87.6}& \textbf{90.5}& \textbf{87.1} & 99.1& \textbf{89.2} & \textbf{77.7} & \textbf{76.9} \\
            \bottomrule
        \end{tabular}
    }
    \vspace{-5mm}
\end{table}

\begin{table}[h]
    \centering
    \caption{\textbf{Imitation Performance of the KungfuBot Student Policy in Isaac Gym.} We evaluate the imitation performance of KungfuBot student policy with unseen motions on the Unitree G1.}
    \label{tab:appendix_D.3_kungfubot_student}
    \scalebox{0.78}{
        \begin{tabular}{l*{10}{c}}
            \toprule
            & \multicolumn{5}{c}{\datasetname \ Test} & \multicolumn{5}{c}{Unseen Video} \\
            \cmidrule(lr){2-6}
            \cmidrule(lr){7-11}
            Dataset & Total & Stationary & Angular & Vertical & Horizontal & Total & Stationary & Angular & Vertical & Horizontal \\
            \midrule
            AMASS & 66.6 & 86.7 & 58.8 & 41.8 & 68.4 & 67.7 & \textbf{97.4} & 68.0 & 59.6 & 37.4 \\
            \datasetname  & \textbf{82.9} & \textbf{93.5}& \textbf{78.5}& \textbf{74.2}& \textbf{81.7}& \textbf{73.8} & \textbf{97.4}& \textbf{76.9} & \textbf{63.8} &  \textbf{47.3} \\
            \bottomrule
        \end{tabular}
    }
    \vspace{-5mm}
\end{table}

\begin{table}[h]
    \centering
    \caption{\textbf{Sim-to-Sim transfer performance across motion dataset.} We evaluate the zero-shot motion tracking success rate of policies trained on \datasetname~and AMASS when transferred from the source simulator (Isaac Gym) to the target simulator (MuJoCo). The results are demonstrated on the G1 humanoid to assess robustness against domain shifts in physics engines.}
    \label{tab:appendix_D.3_sim-to-sim}
    \scalebox{0.78}{
        \begin{tabular}{l*{10}{c}}
            \toprule
            & \multicolumn{5}{c}{\datasetname \ Test} & \multicolumn{5}{c}{Unseen Video} \\
            \cmidrule(lr){2-6}
            \cmidrule(lr){7-11}
            Dataset & Total & Stationary & Angular & Vertical & Horizontal & Total & Stationary & Angular & Vertical & Horizontal \\
            \midrule
            AMASS & 62.1 & 81.4 & 54.2 & 38.8 & 64.3 & 64.3 & 86.2 & 68.5 & 54.3 & 37.4 \\
            \textbf{\datasetname}  & \textbf{75.0} & \textbf{87.6}& \textbf{69.3}& \textbf{61.6}& \textbf{76.3}& \textbf{70.0} & \textbf{87.9}& \textbf{78.8} & \textbf{59.6} & \textbf{38.5} \\
            \bottomrule
        \end{tabular}
    }
    \vspace{-5mm}
\end{table}

\begin{crblock}
\subsection{Evaluation on GMR-Retargeted References}
\label{appendix:gmr_reference}
To test whether the results depend on the retargeting method used for the evaluation references, we retarget the \datasetname~Test and Unseen Video motions with GMR~\citep{araujo2025gmr} and repeat the retargeting and dataset comparisons with the conventional 0.5m success threshold. As shown in Tables~\ref{tab:appendix_cr_gmr_retarget} and~\ref{tab:appendix_cr_gmr_dataset}, PhySINK and \datasetname~achieve the highest success rates on both sets, and \datasetname~also outperforms AMASS retargeted with PhySINK.
\begin{table}[h]
    \centering
    \caption{\textbf{Imitation Performance on GMR-Retargeted References: Retargeting Methods.} Success rate (\%) at the 0.5m threshold of policies trained on AMASS retargeted by each method, evaluated on the \datasetname~Test and Unseen Video motions retargeted with GMR. Mean ($\pm$ std) over three seeds.}
    \label{tab:appendix_cr_gmr_retarget}
    \scalebox{0.85}{
        \begin{tabular}{lcc}
            \toprule
            Retarget & \datasetname~Test (GMR) & Unseen Video (GMR) \\
            \midrule
            IK               & 80.3 ($\pm$0.2) & 81.6 ($\pm$0.6) \\
            GMR              & 80.0 ($\pm$0.1) & 81.1 ($\pm$0.3) \\
            SINK             & 81.6 ($\pm$0.2) & 79.8 ($\pm$0.6) \\
            \textbf{PhySINK} & \textbf{83.4} ($\pm$0.2) & \textbf{83.7} ($\pm$0.4) \\
            \bottomrule
        \end{tabular}
    }
\end{table}

\begin{table}[h]
    \centering
    \caption{\textbf{Imitation Performance on GMR-Retargeted References: Datasets.} Success rate (\%) at the 0.5m threshold of policies trained on each dataset, evaluated on the \datasetname~Test and Unseen Video motions retargeted with GMR. Mean ($\pm$ std) over three seeds.}
    \label{tab:appendix_cr_gmr_dataset}
    \scalebox{0.85}{
        \begin{tabular}{lcc}
            \toprule
            Dataset & \datasetname~Test (GMR) & Unseen Video (GMR) \\
            \midrule
            LaFAN1                & 76.5 ($\pm$0.1) & 81.6 ($\pm$0.9) \\
            Humanoid-X            & 77.7 ($\pm$0.1) & 79.6 ($\pm$0.5) \\
            AMASS (SINK)          & 81.6 ($\pm$0.2) & 79.8 ($\pm$0.6) \\
            AMASS (PhySINK)       & 83.4 ($\pm$0.2) & 83.7 ($\pm$0.4) \\
            \textbf{\datasetname} & \textbf{95.5} ($\pm$0.2) & \textbf{89.4} ($\pm$0.4) \\
            \bottomrule
        \end{tabular}
    }
\end{table}

\subsection{Fall Rate of Humanoid-X and AMASS Policies}
\label{appendix:fall_rate}
We compare how often the policies trained on Humanoid-X and on AMASS fall in simulation, counting an episode as a fall when the pelvis height drops below 0.3m. As shown in Table~\ref{tab:appendix_cr_fall_rate}, the Humanoid-X policy falls about three times as often on both evaluation sets, which poses a safety risk on the real robot. We therefore deploy only the AMASS policy as the real-world baseline.
\begin{table}[h]
    \centering
    \caption{\textbf{Fall Rate in Simulation.} Percentage of episodes in which the pelvis height drops below 0.3m, for policies trained on Humanoid-X and on AMASS retargeted with SINK. Lower is better.}
    \label{tab:appendix_cr_fall_rate}
    \scalebox{0.85}{
        \begin{tabular}{lcc}
            \toprule
            Dataset & \datasetname~Test FR (\%) $\downarrow$ & Unseen Video FR (\%) $\downarrow$ \\
            \midrule
            Humanoid-X   & 18.6 & 13.1 \\
            AMASS (SINK) & \textbf{5.7} & \textbf{4.2} \\
            \bottomrule
        \end{tabular}
    }
\end{table}

\end{crblock}

\section{Ablation Studies}
\label{appendix:ablation_studies_dataset}
\subsection{Mocap only and Video only data performance}
\label{appendix:mocap_vs_video}
To analyze the influence of the human motion source on the downstream motion tracking policy, we divide the PHUMA dataset into two distinct subsets: motions derived from motion capture (Mocap) and motions derived from video-to-motion estimation (Video-Sourced).

We leverage these subsets to train the MaskedMimic policy using identical hyperparameters in Section~\ref{appendix:details_imitation_learning}. As demonstrated in Table~\ref{tab:appendix_E.1_imitation_performance_mocap_vs_video}, the policy trained with video-sourced PHUMA consistently yielded superior imitation performance across all motion categories compared to the policy trained with the mocap-sourced subset.

We attribute this result primarily to the significantly larger and broader motion distribution of the video-sourced data. As illustrated in Figure~\ref{fig:mocap_vs_video}, the video-sourced dataset covers a much broader range of motion types and contains nearly two times more data than the mocap-sourced dataset. Furthermore, as shown in Table~\ref{tab:appendix_E.1_mocap_vs_video}, the PhySINK retargeting method ensures competitive motion quality for both subsets. Because the retargeting quality is similar, the dominant factor leading to the higher imitation performance is the larger size and greater diversity of the video-sourced dataset.
\begin{table}[t!]
    \centering
    \caption{\textbf{Performance Evaluation of PHUMA based on Data Source (MoCap vs. Video).} We present a quantitative comparison evaluating the performance achieved using PHUMA data derived from motion capture (MoCap) versus video, concluding that both sources offer competitive results.}
    \label{tab:appendix_E.1_mocap_vs_video}
    \scalebox{0.80}{
        \begin{tabular}{@{}lcccccccc@{}}
        \toprule
         & Motion Fidelity (\%)  & Joint Feasibility (\%) & Non-Floating (\%) & Non-Penetration (\%) & Non-Skating (\%) \\
        \midrule
        MoCap                                & \textbf{96.7}        & \textbf{100.0}           & \textbf{99.9}          & 94.3          & \textbf{92.1}          \\
        Video                             & 93.8          & \textbf{100.0}           & \textbf{99.9}          & \textbf{98.0}          & 88.3          \\
        \bottomrule
        \end{tabular}
    }  
\end{table}

\begin{table}[t]
    \centering
    \caption{\textbf{Imitation Performance of PHUMA based on Data Source (MoCap vs. Video).} We evaluate the imitation performance of MaskedMimic policy trained with PHUMA data derived
from motion capture (MoCap) versus video on the Unitree G1.}
    \label{tab:appendix_E.1_imitation_performance_mocap_vs_video}
    \scalebox{0.78}{
        \begin{tabular}{l*{10}{c}}
            \toprule
            & \multicolumn{5}{c}{\datasetname \ Test} & \multicolumn{5}{c}{Unseen Video} \\
            \cmidrule(lr){2-6}
            \cmidrule(lr){7-11}
            Dataset & Total & Stationary & Angular & Vertical & Horizontal & Total & Stationary & Angular & Vertical & Horizontal \\
            \midrule
            MoCap & 75.2 & 88.2 & 68.4 & 49.0 & 86.9 & 73.0 & 96.6 & 73.9 & 56.4 & 58.2 \\
            Video  & \textbf{85.7} & \textbf{92.9}& \textbf{81.8}& \textbf{75.5}& \textbf{89.5}& \textbf{76.2} & \textbf{100.0}& \textbf{76.4} & \textbf{59.6} & \textbf{62.6} \\
            \bottomrule
        \end{tabular}
    }
\end{table}

\begin{crblock}
\textbf{Adding video to a mocap base.} We start from the mocap-sourced motions and add 33\%, 66\%, and 100\% of the video-sourced motions, retargeting all of them with PhySINK. As shown in Table~\ref{tab:appendix_cr_data_source}, performance on both evaluation sets is highest with the full video set.
\begin{table}[h]
    \centering
    \caption{\textbf{Imitation Performance When Adding Video to Mocap.} Success rate (\%) at the 0.5m threshold on GMR-retargeted references, for policies trained on the mocap-sourced motions plus a growing fraction of the video-sourced motions. Mean ($\pm$ std) over three seeds.}
    \label{tab:appendix_cr_data_source}
    \scalebox{0.85}{
        \begin{tabular}{lcc}
            \toprule
            Data Source & \datasetname~Test (GMR) & Unseen Video (GMR) \\
            \midrule
            MoCap Only        & 82.2 ($\pm$0.2) & 86.0 ($\pm$0.2) \\
            $+$ Video (33\%)  & 82.1 ($\pm$0.1) & 80.1 ($\pm$0.8) \\
            $+$ Video (66\%)  & 92.8 ($\pm$0.1) & 86.7 ($\pm$0.5) \\
            $+$ Video (100\%) & \textbf{94.6} ($\pm$0.1) & \textbf{90.2} ($\pm$0.9) \\
            \bottomrule
        \end{tabular}
    }
\end{table}

\end{crblock}

\subsection{Physics-based Filtering}
\label{appendix:dataset_curation}
This section provides ablation studies on the physics-based filtering criteria used in data curation (Table~\ref{tab:appendix_A.1_filtering_criteria}) and the \crnew{physics-aware} losses used in PhySINK.

\subsubsection{Data Distribution Based on Physics-based Filtering}
\label{appendix:data_distribution_physics_filtering}
\begin{figure}[htbp]
    \centering
    \includegraphics[width=0.9\textwidth]{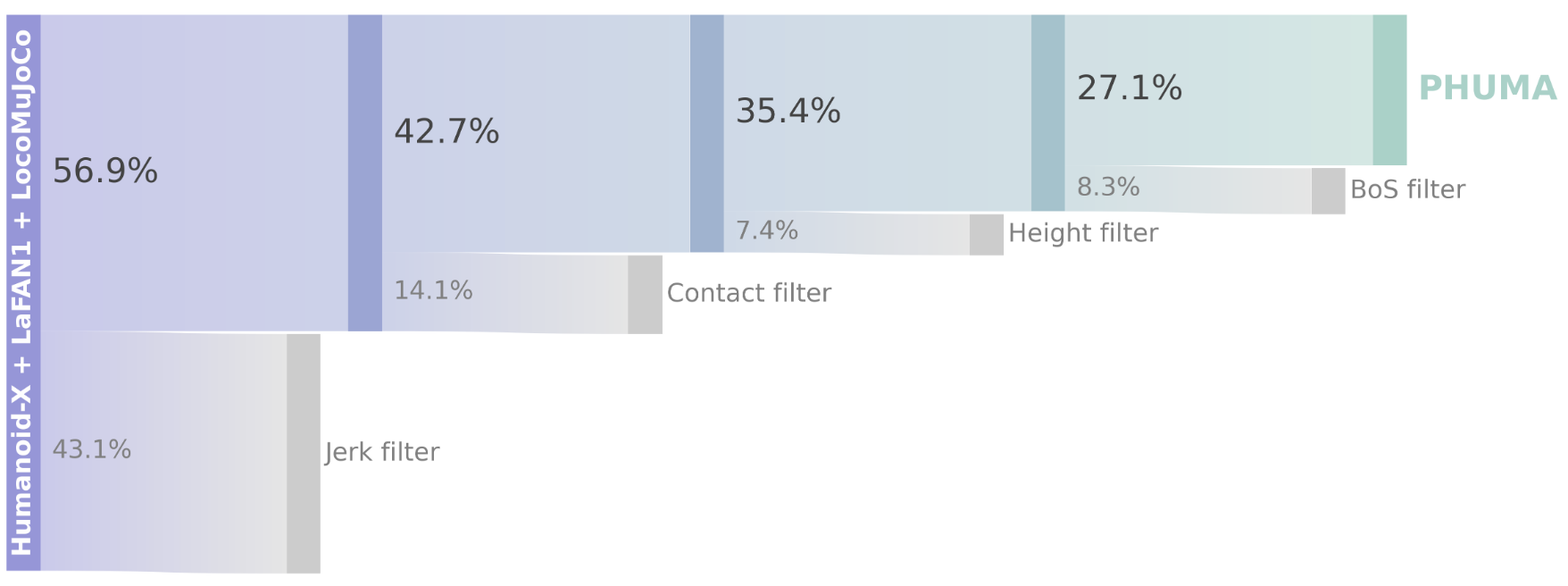}
    \caption{\textbf{Dataset Statistics After Physics-based Filtering.} Distribution of motion sequences after applying physics-based filtering to the combined Humanoid-X, LaFAN1, and LocoMuJoCo datasets.}
    \label{fig:appendix_E.2_curation_filtering_statistics}
\end{figure}

Figure~\ref{fig:appendix_E.2_curation_filtering_statistics} shows the filtering statistics when sequentially applying the physics-based criteria from Table~\ref{tab:appendix_A.1_filtering_criteria} to Humanoid-X~\citep{mao2025humanoidx}. The filters are applied in the following order: (1) root jerk filter (jerk $<50\,\text{m/s}^3$), (2) contact filter (foot contact score $>0.6$), (3) height filter (minimum pelvis height $>0.6\,\text{m}$ and maximum pelvis height $<1.5\,\text{m}$), and (4) base of support (BoS) filter (pelvis distance to BoS $<6\,\text{cm}$ and spine1 distance to BoS $<11\,\text{cm}$). After applying all filters sequentially, 27.1\% of the original Humanoid-X dataset remains, representing motions that satisfy physical plausibility constraints.

\subsubsection{PhySINK's Robustness to Noisy Motion Sources}
\label{appendix:physink_robustness}
To evaluate how robustly PhySINK handles noisy human motion inputs, we retarget Humanoid-X motion sources with varying levels of filtering: (1) raw Humanoid-X (no filtering), (2) Humanoid-X + jerk filtering, (3) Humanoid-X + foot contact filtering, (4) Humanoid-X + height filtering, (5) Humanoid-X + BoS filtering, and (6) Humanoid-X + all filters. We then apply PhySINK to each variant. Note that we exclude pre-retargeted datasets (LaFAN1 and LocoMuJoCo) from this analysis to isolate the effect of filtering on Humanoid-X.
As shown in Table~\ref{tab:appendix_E.2_physink_based_on_noisy_motion}, PhySINK demonstrates robust retargeting performance across motion sources with varying noise levels, successfully handling physical implausibilities present in the raw data.

\begin{table}[t!]
    \centering
    \caption{\textbf{PhySINK Retargeting Robustness to Noisy Motion Sources.} This figure presents an experiment to evaluate how robustly the PhySINK method retargets various noisy Humanoid-X motion sources. The six distinct motion source groups used for retargeting are compared: (1) Original Humanoid-X, and Humanoid-X motion sources sequentially refined by applying (2) Jerk filtering, (3) Foot Contact filtering, (4) Height filtering, (5) BoS filtering, and (6) All filtering (PHUMA).}
    \label{tab:appendix_E.2_physink_based_on_noisy_motion}
    \scalebox{0.70}{
        \begin{tabular}{@{}lccccccccc@{}}
        \toprule
         Motion Source & Hours & Motion Fidelity (\%)  & Joint Feasibility (\%) & Non-Floating (\%) & Non-Penetration (\%) & Non-Skating (\%) \\
        \midrule
        Humanoid-X                       &  237.2         & 70.6     & \textbf{100.0}           & 98.7          & 92.2          & 90.1          \\
        Jerk Filter                     &  141.1        & 76.9        & \textbf{100.0}          & 99.7          & 96.5          & \textbf{92.0}          \\
        Foot Contact Filter             &  123.1        & 77.1        & \textbf{100.0}          & 99.8          & \textbf{96.8}          & 90.6          \\
        Height Filter                    &  135.7       & 87.0        & \textbf{100.0}          & 99.4          & 94.7          & 89.8          \\
        BoS Filter                       &  110.3      & 90.7        & \textbf{100.0}         & 99.4          & 94.9          & 90.0          \\
        All Filter                       &   62.2       & \textbf{94.8}      & \textbf{100.0}           & \textbf{99.9}          & 96.7          & 89.7          \\
        \bottomrule
        \end{tabular}
    }  
\end{table}

\begin{crblock}
\subsubsection{Leave-One-Out Ablation of Curation Filters}
\label{appendix:loo_curation}
We remove one filter at a time from the full curation pipeline, retarget the remaining motions with PhySINK, and train a policy on each variant. As shown in Table~\ref{tab:appendix_cr_loo_curation}, applying all filters yields the best performance, and removing the foot contact filter causes the largest drop.
\begin{table}[h]
    \centering
    \caption{\textbf{Leave-One-Out Ablation of Curation Filters.} Success rate (\%) at the 0.5m threshold on GMR-retargeted references, for policies trained on motions curated with all filters or with one filter removed, all retargeted with PhySINK. Hours give the curated volume. Mean ($\pm$ std) over three seeds.}
    \label{tab:appendix_cr_loo_curation}
    \scalebox{0.85}{
        \begin{tabular}{lccc}
            \toprule
            Curation Filter & Hours & \datasetname~Test (GMR) & Unseen Video (GMR) \\
            \midrule
            \textbf{All Filters} & 62 & \textbf{94.6} ($\pm$0.1) & \textbf{90.2} ($\pm$0.9) \\
            No Jerk Filter       & 75 & 91.8 ($\pm$0.1) & 89.5 ($\pm$0.3) \\
            No FC Filter         & 77 & 87.5 ($\pm$0.1) & 80.0 ($\pm$0.5) \\
            No Height Filter     & 73 & 89.2 ($\pm$0.3) & 85.6 ($\pm$0.1) \\
            No BoS Filter        & 88 & 89.5 ($\pm$0.1) & 87.4 ($\pm$0.7) \\
            \bottomrule
        \end{tabular}
    }
\end{table}

\subsubsection{Sensitivity to the Video Reconstruction Model}
\label{appendix:recon}
We reconstruct the Unseen Video motions with TRAM~\citep{wang2024tram}, used in \datasetname, and with GVHMR~\citep{shen2024gvhmr}, then curate them and retarget the accepted motions with SINK and PhySINK. As shown in Table~\ref{tab:appendix_cr_recon}, GVHMR yields a higher curation acceptance rate, yet the better reconstruction does not close the quality gap between SINK and PhySINK.
\begin{table}[h]
    \centering
    \caption{\textbf{Sensitivity to the Video Reconstruction Model.} Curation acceptance rate on the Unseen Video motions and retargeting quality of SINK and PhySINK, for motions reconstructed with TRAM and GVHMR.}
    \label{tab:appendix_cr_recon}
    \scalebox{0.8}{
        \begin{tabular}{l|c|l|ccccc}
            \toprule
            \multirow{2}{*}{Reconstruction} & Acceptance & \multirow{2}{*}{Retarget} & \multicolumn{5}{c}{Retargeting Quality} \\
            \cmidrule(lr){4-8}
            & Rate (\%) & & MF (\%) & JF (\%) & NF (\%) & NP (\%) & NS (\%) \\
            \midrule
            \multirow{2}{*}{TRAM}  & \multirow{2}{*}{41.9}
                & SINK    & \textbf{95.9} & 63.6 & \textbf{100.0} & 27.2 & 40.1 \\
                & & PhySINK & \textbf{95.9} & \textbf{100.0} & \textbf{100.0} & \textbf{99.7} & \textbf{93.8} \\
            \midrule
            \multirow{2}{*}{GVHMR} & \multirow{2}{*}{\textbf{77.4}}
                & SINK    & \textbf{95.6} & 78.7 & 99.9 & 27.3 & 64.8 \\
                & & PhySINK & \textbf{95.6} & \textbf{100.0} & \textbf{100.0} & \textbf{100.0} & \textbf{86.6} \\
            \bottomrule
        \end{tabular}
    }
\end{table}

\end{crblock}

\subsection{Impact of Motion Retargeting Quality on Policy Performance}
\label{appendix:dataset_artifact}
To investigate how physical artifacts in retargeted motion data affect policy learning, we train MaskedMimic policies on datasets generated using six retargeting methods with varying artifact levels. The methods are: (1) \textbf{IK}, which produces significant artifacts in motion fidelity, joint limits, grounding, and skating; (2) \textbf{GMR}, which reduces motion fidelity loss, grounding issues, and skating compared to IK; (3) \textbf{SINK}, which improves motion fidelity and joint limit violations; (4) \textbf{SINK + Joint Feasibility Loss}, which further reduces joint limit violations; (5) \textbf{SINK + Joint Feasibility + Grounding Loss}, which addresses all artifact types except skating; and (6) \textbf{PhySINK}, which minimizes all physical artifacts. To isolate the effect of retargeting quality, we exclude the LaFAN1 and LocoMuJoCo datasets from this analysis, as they were pre-retargeted and would not allow for fair comparison across methods.

In Table~\ref{tab:appendix_E.3_imitation_performance_retargeting_loss}, our results show that SINK-based methods, which first optimize the humanoid body shape before applying it to the original motion, consistently outperform IK-based methods that rely on heuristic scaling to bridge human-humanoid discrepancies. Notably, GMR achieves better performance than IK despite having similar joint limit issues, while the SINK variants further outperform GMR in tracking performance. However, across the SINK variants themselves (SINK, SINK + Joint Feasibility Loss, SINK + Joint Feasibility + Grounding Loss), we observe comparable performance despite their similar levels of motion fidelity, joint feasibility, and grounding quality, the main differences among these variants being penetration and skating artifacts. 

These results indicate that motion fidelity is the most critical factor affecting motion tracking performance. The substantial performance improvement of GMR over IK (achieved primarily through motion fidelity gains rather than joint feasibility improvements) demonstrates that preserving motion fidelity during retargeting has a greater impact on policy learning than other artifact types. However, once motion fidelity reaches a certain threshold (as in the SINK variants), further reductions in other artifact types yield diminishing returns for policy learning.
\begin{table}[h]
    \centering
   \caption{\textbf{Ablation Studies of Imitation Performance on Retargeting Loss.} We evaluate the imitation performance of MaskedMimic policies trained with and without the physical constraint loss (Table~\ref{tab:retarget_ablation}) using the Unitree G1 robot.}
    \label{tab:appendix_E.3_imitation_performance_retargeting_loss}
    \scalebox{0.72}{
        \begin{tabular}{l*{10}{c}}
            \toprule
            & \multicolumn{5}{c}{\datasetname \ Test} & \multicolumn{5}{c}{Unseen Video} \\
            \cmidrule(lr){2-6}
            \cmidrule(lr){7-11}
            Dataset & Total & Stationary & Angular & Vertical & Horizontal & Total & Stationary & Angular & Vertical & Horizontal \\
            \midrule
            IK & 70.5 & 85.9 & 63.3 & 47.6 & 77.3 & 68.5 & 96.6 & 72.4 & 43.6 & 49.5 \\
            GMR & 84.0 & 92.1 & 77.8 & 77.1 & 89.1 & 75.2 & 99.1 & 77.8 & 61.7 & 52.7 \\
            SINK & 89.1 & 94.0 & 86.0 & 84.9 & 90.7 & 79.0 & \textbf{100.0} & 81.8 & \textbf{62.8} & 62.6 \\
            + Joint Feasibility Loss & 87.0 & 92.1 & 83.6 & 79.8 & 90.9 & 78.6 & 94.8 & 84.7 & 56.4 & 67.0 \\
            + Grounding Loss & \textbf{90.0} & 93.5 & \textbf{87.9} & \textbf{85.6} & \textbf{92.0} & 80.4 & 98.3 & 86.2 & 60.6 & 64.8 \\
            + Skating Loss & 89.9 & \textbf{94.2} & 87.6 & 84.2 & 91.8 & \textbf{81.7} & 97.4 & \textbf{86.7} & 61.7 & \textbf{71.4} \\
            \bottomrule
        \end{tabular}
    }
\end{table}

\begin{crblock}
\subsection{Leave-One-Out Ablation of PhySINK Losses}
\label{appendix:loo_retarget}
We remove one loss at a time from the full PhySINK objective and train a policy on each variant. As shown in Table~\ref{tab:appendix_cr_loo_retarget}, removing a loss degrades the retargeting metric that loss governs, while the full objective achieves the highest tracking performance. Removing the grounding loss causes the largest drop.
\begin{table}[h]
    \centering
    \caption{\textbf{Leave-One-Out Ablation of PhySINK Losses.} Retargeting quality, and success rate (\%) at the 0.5m threshold on GMR-retargeted references, with each loss removed from the full objective. Mean ($\pm$ std) over three seeds.}
    \label{tab:appendix_cr_loo_retarget}
    \resizebox{\columnwidth}{!}{
        \begin{tabular}{l|ccccc|cc}
            \toprule
            \multirow{2}{*}{Retarget} & \multicolumn{5}{c|}{Retargeting Quality} & \multicolumn{2}{c}{Motion Tracking Performance} \\
            \cmidrule(lr){2-6} \cmidrule(lr){7-8}
            & MF (\%) & JF (\%) & NF (\%) & NP (\%) & NS (\%) & \datasetname~Test (GMR) & Unseen Video (GMR) \\
            \midrule
            \textbf{PhySINK}          & 94.8 & \textbf{100.0} & \textbf{99.9} & 96.8 & 89.7 & \textbf{94.6} ($\pm$0.1) & \textbf{90.2} ($\pm$0.9) \\
            No Joint Feasibility Loss & 91.7 & 95.8 & \textbf{99.9} & 96.4 & 89.4 & 91.0 ($\pm$0.1) & 85.8 ($\pm$0.6) \\
            No Grounding Loss         & 91.8 & \textbf{100.0} & 96.4 & 14.6 & \textbf{91.7} & 90.5 ($\pm$0.1) & 84.5 ($\pm$0.8) \\
            No Skating Loss           & \textbf{94.9} & \textbf{100.0} & \textbf{99.9} & \textbf{97.2} & 53.6 & 94.2 ($\pm$0.1) & 88.2 ($\pm$0.8) \\
            \bottomrule
        \end{tabular}
    }
\end{table}

\end{crblock}

\section{Pelvis-Only Path Following Control Performance}
\label{appendix:path_following}

We evaluate whether training on \datasetname~enables better pelvis path-following control compared to the AMASS dataset. Using MaskedMimic's partially-constrained protocol, we train two student policies: one distilled from an AMASS-trained teacher and another from a \datasetname-trained teacher. Both students receive only pelvis position and rotation as input.

\begin{figure}[h]
    \centering
    \includegraphics[width=0.9\textwidth]{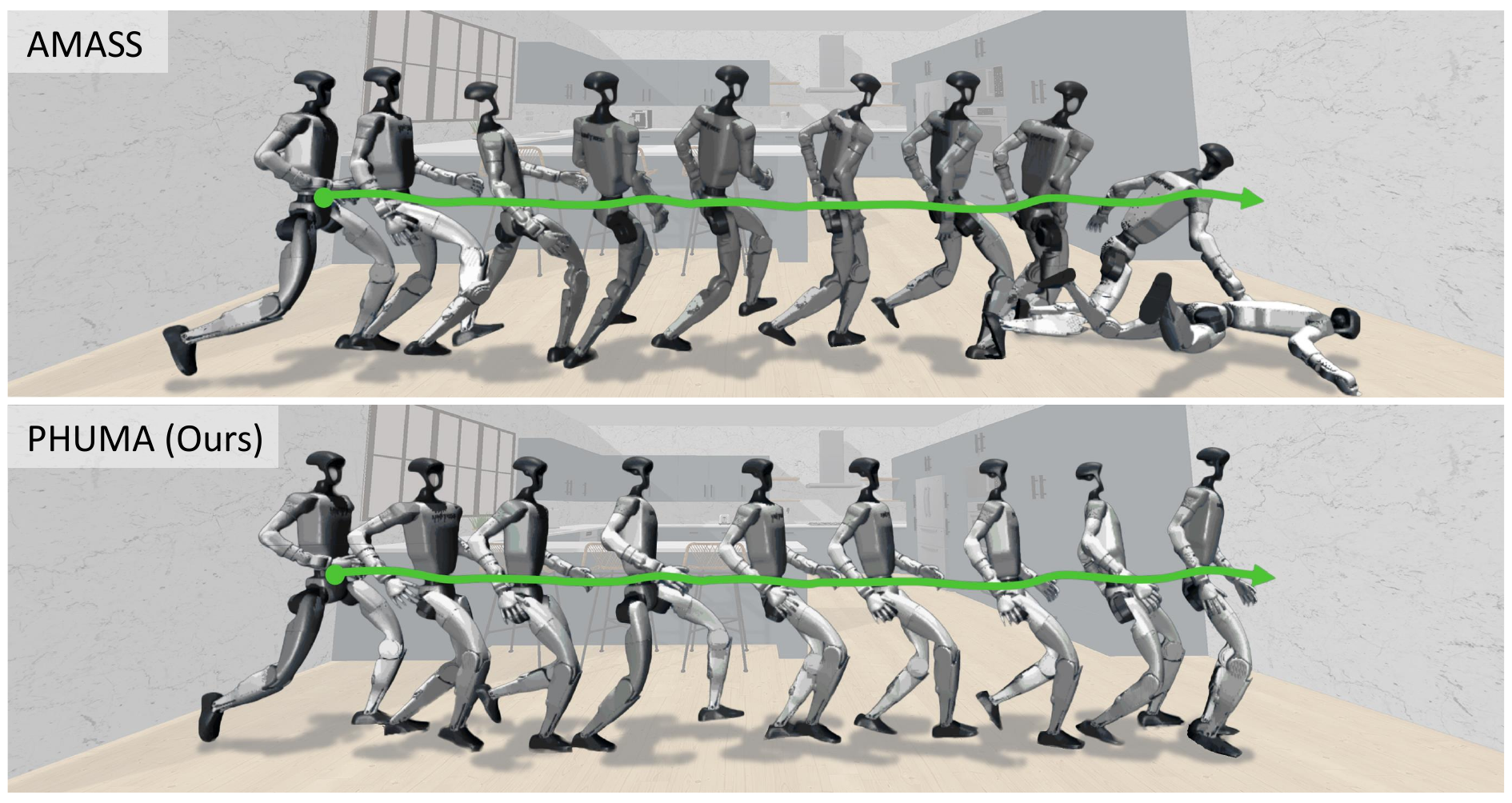}
    \vspace{-1mm}
    \caption{\textbf{Path following on running motion.} We visualize the robot's trajectory in a running motion. The target pelvis path is visualized with a green line. Top row presents results from a policy trained on AMASS, while bottom row presents results from a policy trained on \datasetname. }
    \label{fig:appendix_F_path_following}
\end{figure}

As shown in Table~\ref{tab:appendix_F_path_following}, policies trained on \datasetname~outperform those trained on baseline datasets across all motion categories and humanoids, with the largest gap on dynamic motions. Figure~\ref{fig:appendix_F_path_following} illustrates this, where AMASS-trained policies fail on running while \datasetname-trained policies remain robust.
\begin{table}[h]
    \centering
    \vspace{-4mm}
    \caption{\textbf{Pelvis path following peformance across motion dataset.} We evaluate the success rate of pelvis path-following control for policies trained on the AMASS and PHUMA datasets across various pelvis trajectories from the PHUMA Test and Unseen Video.}
    \vspace{1mm}
    \label{tab:appendix_F_path_following}
    \scalebox{0.78}{
        \begin{tabular}{l*{10}{c}}
            \toprule
            & \multicolumn{5}{c}{\datasetname \ Test} & \multicolumn{5}{c}{Unseen Video} \\
            \cmidrule(lr){2-6}
            \cmidrule(lr){7-11}
            Dataset & Total & Stationary & Angular & Vertical & Horizontal & Total & Stationary & Angular & Vertical & Horizontal \\
            \midrule
            \rowcolor{header}\multicolumn{11}{l}{\textbf{(a) G1}} \\
            \midrule
            AMASS & 60.5 & 85.6 & 60.1 & 51.4 & 66.5 & 54.8 & 83.6 & 66.5 & 33.0 & 27.5 \\
            \textbf{\datasetname}  & \textbf{84.5} & \textbf{94.6}& \textbf{86.1}& \textbf{83.7}& \textbf{90.2}& \textbf{74.6} & \textbf{98.3}& \textbf{83.3} & \textbf{54.3} & \textbf{57.1} \\
            \midrule
            \rowcolor{header}\multicolumn{11}{l}{\textbf{(a) H1-2}} \\
            \midrule
            AMASS & 60.4 & 84.0& 62.8& 43.6& 78.7& 72.3 & \textbf{96.6} & 77.3 & 52.1 & 72.5  \\
            \textbf{\datasetname} & \textbf{73.9}& \textbf{91.2}& \textbf{76.5}& \textbf{66.9}& \textbf{84.8}& \textbf{78.1} & \textbf{96.6} & \textbf{77.8} & \textbf{60.6} & \textbf{78.0}  \\
            \bottomrule
        \end{tabular}
    }
\end{table}

\section{Results on H1-2 Robot}
\label{appendix:h12_results}

This appendix provides complete results on the Unitree H1-2~\citep{UnitreeH12} robot, complementing the G1 results presented in the main text.

\subsection{PhySINK Ablation on H1-2}
Table~\ref{tab:retarget_ablation_h12} presents the ablation study of PhySINK retargeting on the H1-2 humanoid robot, showing the progressive impact of each physical constraint loss.

\begin{table}[h]
    \centering
    \caption{\textbf{Quantitative comparison and ablation study of retargeting methods on H1-2.} Progressive impact of adding each physical constraint loss.}
    \label{tab:retarget_ablation_h12}
    \scalebox{0.71}{
        \begin{tabular}{@{}lcccccccc@{}}
        \toprule
         & Motion Fidelity (\%)  & Joint Feasibility (\%) & Non-Floating (\%) & Non-Penetration (\%) & Non-Skating (\%) \\
        \midrule
        IK               & 36.3          & 80.9           & 57.7          & 45.2          & 56.1          \\
        GMR & 58.0 & 71.8 & 6.0 & \textbf{100.0} & 62.3 \\
        SINK & 93.9 & 15.3 & 42.2 & 81.4 & 47.9 \\
        + Joint Feasibility Loss          & \textbf{94.0} & 99.5           & 44.4          & 79.9          & 50.7          \\
        + Grounding Loss          & 93.9          & \textbf{99.9} & \textbf{99.8}          & 98.1          & 49.3          \\
        + Skating Loss = PhySINK          & 93.9          & \textbf{99.9} & 97.7          & 99.7          & \textbf{87.7}          \\
        \bottomrule
        \end{tabular}
    }
\end{table}

\subsection{Retargeting Method Comparison on H1-2}
Table~\ref{tab:tracking_comparison_retarget_h12} shows motion tracking performance across retargeting approaches on the H1-2 robot.

\begin{table}[h]
    \centering
    \caption{\textbf{Motion tracking performance across retargeting approaches on H1-2.} Success rates using two test sets.}
    \label{tab:tracking_comparison_retarget_h12}
    \scalebox{0.77}{
        \begin{tabular}{l*{11}{c}}
            \toprule
            & & \multicolumn{5}{c}{\datasetname \ Test} & \multicolumn{5}{c}{Unseen Video} \\
            \cmidrule(lr){2-6}
            \cmidrule(lr){7-11}
            Retarget  & Total & Stationary & Angular & Vertical & Horizontal & Total & Stationary & Angular & Vertical & Horizontal  \\
            \midrule
            IK               & 45.3 & 70.9 & 35.7 & 15.2 & 35.0 & 54.2 & 78.0 & 60.7 & 30.1 & 28.6 \\
            GMR              & 56.6 & 75.7 & 48.1 & 24.8 & \textbf{65.9} & 64.3 & 80.2 & 65.5 & 52.1 & 53.8 \\
            SINK             & 54.4 & 74.9 & 45.9 & 17.2 & 49.6 & 64.3 & 87.3 & 59.7 & 46.0 & \textbf{63.9} \\
            \textbf{PhySINK} & \textbf{64.3} & \textbf{83.6} & \textbf{57.0} & \textbf{27.7} & 55.9 & \textbf{72.4} & \textbf{99.2} & \textbf{66.3} & \textbf{57.4} & 63.1 \\
            \bottomrule
        \end{tabular}
    }
\end{table}

\subsection{Dataset Effectiveness on H1-2}
Table~\ref{tab:tracking_comparison_dataset_h12} presents motion tracking performance across training datasets on the H1-2 robot.

\begin{table}[h]
    \centering
    \caption{\textbf{Motion tracking performance across datasets on H1-2.} Success rates evaluated on two test sets.}
    \label{tab:tracking_comparison_dataset_h12}
    \scalebox{0.7}{
        \begin{tabular}{l*{11}{c}}
            \toprule
            & & \multicolumn{5}{c}{\datasetname \ Test} & \multicolumn{5}{c}{Unseen Video} \\
            \cmidrule(lr){3-7}
            \cmidrule(lr){8-12}
            Dataset & Hours & Total & Stationary & Angular & Vertical & Horizontal & Total & Stationary & Angular & Vertical & Horizontal  \\
            \midrule
            LaFAN1                & 2.4   & 62.0 & 79.3 & 54.7 & 26.6 & 58.9 & 70.8 & 92.4 & 66.7 & 56.4 & \textbf{68.2} \\
            AMASS                 & 20.9 & 54.4 & 74.9 & 45.9 & 17.2 & 49.6 & 64.3 & 87.3 & 59.7 & 46.0 & 63.9 \\
            Humanoid-X            & 231.4 & 49.7 & 74.6 & 40.4 & 17.0 & 37.3 & 60.5 & 88.3 & 60.0 & 48.7 & 39.7 \\
            \textbf{\datasetname} & 73.0  & \textbf{82.7} & \textbf{91.5} & \textbf{79.5} & \textbf{68.1} & \textbf{68.4} & \textbf{78.6} & \textbf{97.5} & \textbf{76.8} & \textbf{74.5} & 63.8 \\
            \bottomrule
        \end{tabular}
    }
\end{table}

\end{document}